\documentclass[10pt]{elsarticle}

\usepackage{amssymb}
\usepackage[tight,footnotesize]{subfigure}
\usepackage[cmex10]{amsmath}
\usepackage{url}

\newcommand{\vect}[1]{ \boldsymbol{#1} }
\newcommand{\matr}[1]{ \boldsymbol{#1} }

\renewcommand{\Re}{\text{Re}}
\renewcommand{\Im}{\text{Im}}

\newcommand{\CGaussPDF}{\mathrm{CN}}
\newcommand{\GaussPDF}{\mathrm{N}}
\newcommand{\GammaPDF}{\mathrm{Ga}}

\newcommand{\compnum}{\mathbb{C}}
\newcommand{\realnum}{\mathbb{R}}
\newcommand{\w}{\vect{w}}
\newcommand{\n}{\vect{n}}
\newcommand{\what}{\hat{\w}}
\newcommand{\gam}{\vect{\gamma}}
\newcommand{\gamhat}{\hat{\gam}}
\newcommand{\gammahat}{\hat{\gamma}}
\newcommand{\gammatilde}{\tilde{\gamma}}

\newcommand{\muv}{\vect{\mu}}
\newcommand{\muvhat}{\hat{\muv}}
\newcommand{\lambdahat}{\hat{\lambda}}
\newcommand{\y}{\vect{y}}
\newcommand{\Hmat}{\matr{\Phi}}
\newcommand{\Cmat}{\matr{C}}
\newcommand{\Imat}{\matr{I}}
\newcommand{\Gammat}{\matr{\Gamma}}
\newcommand{\Gammathat}{\widehat{\Gammat}}
\newcommand{\Sigmamat}{\matr{\Sigma}}

\newcommand{\h}{\vect{\phi}}
\newcommand{\ev}{\vect{e}}
\newcommand{\supp}{\mathrm{supp}}

\newcommand{\ti}{Type I}
\newcommand{\tii}{Type II}

\newcommand{\Fig}{Fig.}

\DeclareMathOperator*{\argmax}{argmax}
\DeclareMathOperator*{\diag}{diag}

\newcommand{\trans}{\mathrm{T}}
\newcommand{\hermit}{\mathrm{H}}

\usepackage{lipsum}
\makeatletter
\def\ps@pprintTitle{%
 \let\@oddhead\@empty
 \let\@evenhead\@empty
 \def\@oddfoot{}%
 \let\@evenfoot\@oddfoot}
\makeatother

\begin{document}

\begin{frontmatter}



\title{Sparse Estimation Using Bayesian Hierarchical Prior Modeling for Real and Complex Linear Models\tnoteref{label_title}}
\tnotetext[label_title]{Parts of this work have previously been presented in conference proceedings \cite{Pedersen2012,Pedersen2013}.}


\author[label1]{Niels Lovmand Pedersen\corref{cor1}}
\ead{nlp@es.aau.dk}
\author[label1]{Carles Navarro Manch\'{o}n}
\ead{cnm@es.aau.dk}
\author[label1]{Mihai-Alin Badiu}
\ead{in\_mba@es.aau.dk}
\author[label2]{Dmitriy Shutin}
\ead{dmitriy.shutin@dlr.de}
\author[label1]{Bernard Henri Fleury}
\ead{bfl@es.aau.dk}

\address[label1]{Department of Electronic Systems, Aalborg University, 
Niels Jernes Vej 12,\\DK-9220 Aalborg, Denmark.}
\address[label2]{Institute of Communications and Navigation, German Aerospace Center,
Oberpfaffenhofen, D-82234 Wessling, Germany.}

\begin{abstract}

In sparse Bayesian learning (SBL), Gaussian scale mixtures (GSMs) have been used to model sparsity-inducing priors that realize a class of concave penalty functions for the regression task in real-valued signal models. Motivated by the relative scarcity of formal tools for SBL in complex-valued models, this paper proposes a GSM model -– the Bessel K model –- that induces concave penalty functions for the estimation of complex sparse signals. The properties of the Bessel K model are analyzed when it is applied to Type I and Type II estimation. This analysis reveals that, by tuning the parameters of the mixing pdf different penalty functions are invoked depending on the estimation type used, the value of the noise variance, and whether real or complex signals are estimated. Using the Bessel K model, we derive a sparse estimator based on a modification of the expectation-maximization algorithm formulated for Type II estimation. The estimator includes as a special instance the algorithms proposed by Tipping and Faul \cite{Tipping2003} and by Babacan \emph{et al.} \cite{Babacan2010}. Numerical results show the superiority of the proposed estimator over these state-of-the-art estimators in terms of convergence speed, sparseness, reconstruction error, and robustness in low and medium signal-to-noise ratio regimes.

\end{abstract}


\end{frontmatter}


\section{Introduction \label{sec:introduction}}
Compressive sensing and sparse signal representation have attracted the interest of an increasing number of researchers over the recent years \cite{BaraniukTutorial2007,Wakin2008,Tzikas2008,Bajwa2010}. This is motivated by the widespread applicability that such techniques have found in a large variety of engineering disciplines.  
Generally speaking, these disciplines consider the following signal model:
\begin{align}
	\y=\Hmat\w+\n.
	\label{eq:model}
\end{align}
In this expression, $\y$ is an $M\times 1$ vector of measurement samples, $\Hmat=[\h_1,\ldots,\h_N]$ is an $M\times N$ dictionary matrix with $N > M$. The additive term $\n$ is an $M\times 1$ perturbation vector, which is assumed to be Gaussian distributed with zero-mean and covariance $\lambda^{-1}\matr{I}$, where $\lambda > 0$ denotes the noise precision and $\matr{I}$ is the identity matrix. The objective is to accurately estimate the $N\times1$ unknown weight vector $\w = [w_1,\ldots,w_N]^\trans$, which is assumed $K$-sparse in the canonical basis.  

We coin the signal model \eqref{eq:model} as either real, when $\Hmat$, $\w$, and $\n$ are all real, or as complex, when $\Hmat$, $\w$, and $\n$ are all complex.\footnote{Obviously, one could also consider a mixed model where, e.g., $\Hmat$ and $\n$ are complex but $\w$ is real. In this paper we focus on the two most relevant cases of real and complex signal models as defined above.} Historically, real signal models have dominated the research in sparse signal representation and compressive sensing. However, applications seeking sparse estimation for complex signal models are not uncommon. An example is the estimation of multipath wireless channels \cite{Bajwa2010,ShutinFleuryVBSAGEChannel,Pedersen2012,Pedersen2013}. The extension of sparse representation from real signal models to complex models is not always straightforward, as we will discuss in this paper. 

Many convex \cite{Tibshirani1994,Chen1998}, greedy \cite{Tropp2004,Needell2009}, and Bayesian methods have been proposed in the literature in recent years to devise sparse estimators. In this paper, we focus on Bayesian inference methods commonly referred to as sparse Bayesian learning (SBL) \cite{Tipping2001, WipfRao04}. In SBL, we design priors for $\w$ that induce sparse representations of $\Hmat\w$. Instead of working directly with the prior probability density function (pdf) $p(\w)$, SBL typically uses a two-layer hierarchical prior model that involves a conditional prior pdf $p(\w|\gam)$ and a hyperprior pdf $p(\gam)$. The goal is to select these pdfs in such a way that we can construct  computationally tractable iterative algorithms that estimate both the hyperparameter vector $\gam$ and the weight vector $\w$ with the latter estimate being sparse. A widely used two-layer prior model is the model where the entries of $\w$ are independent and identically distributed according to a Gaussian scale mixture (GSM) \cite{Andrews1974,Barndorff-Nielsen1982,Gneiting1997,Eltoft2006,Palmer2006}. Specifically, $w_i$ is modeled as $w_i = \sqrt{\gamma_i}u_i$, where $u_i$ is a standard Gaussian random variable and $\gamma_i$ is a nonnegative random scaling factor, also known as the mixing variable.\footnote{In this configuration, $\gamma_i$ can be seen as the variance of $w_i$.} The pdf $p(\gamma_i)$ of the latter variable is called the mixing pdf of the GSM. Based on a careful selection of $p(\gamma_i)$ an inference algorithm is then constructed. The sparsity-inducing property of the resulting estimator does not only depend on $p(\gamma_i)$ but also on the type of inference method used, as discussed next.     

In SBL two widespread inference approaches, referred to as \textit{\ti} and \textit{\tii{}} estimation following \cite{Wipf2011}, have been used. 
In \ti{} estimation, the maximum-a-posteriori (MAP) estimate of $\w$ is computed from the observation $\y$:
\begin{align}
		\what_{I}(\y) &= \argmax_{\w} p(\w|\y)  \notag\\
		&= \argmax_{\w} \log \int p(\y|\w)p(\w|\gam)p(\gam)\mathrm{d}\gam.
		\label{eq:wmap}
\end{align}
Equivalently, the \ti{} estimator $\what_I$ is obtained as the minimizer of the \ti{} cost function 
\begin{align}
\mathcal{L}_{I}(\w) \triangleq \rho\|\y-\Hmat\w\|^2_2+\lambda^{-1}q_{I}(\w).
\label{eq:type1cost}
\end{align} 
In the above expression, $\|\cdot\|_p$, $p\geq1$, is the $\ell_p$-norm and the parameter $\rho$ takes values $\rho=1/2$ when the signal model \eqref{eq:model} is real and $\rho=1$ when it is complex. The pdf $p(\gam)$ is designed such that the penalization term $q_{I}(\w) \propto^{e} -\log p(\w)$ with $p(\w) = \int p(\w|\gam) p(\gam) \mathrm{d}\gam$ enforces a sparse estimate of the weight vector $\w$.\footnote{Here $x\propto^{e}y$ denotes $\exp(x) = \exp(\upsilon) \exp(y)$, and thus $x = \upsilon + y$, for some arbitrary constant $\upsilon$. We will also make use of $x\propto y$, which denotes $x = \upsilon y$ for some positive constant $\upsilon$.} 

In \tii{} estimation \cite{MacKay1992,Tipping2001,WipfRao04}, the MAP estimate of $\gam$ is computed from the observation $\y$: 
\begin{align}
\gamhat_{II}(\y) &= \argmax_{\gam} p(\gam|\y)  \notag\\
&= \argmax_{\gam} \log \int p(\y|\w)p(\w|\gam)p(\gam)\mathrm{d}\w.
\label{eq:type2gam}
\end{align}
Thus, the estimator $\gamhat_{II}$ is the minimizer of 
\begin{align}
\mathcal{L}_{II}(\gam)  \triangleq \rho\y^\hermit\Cmat^{-1}\y + \rho\log|\Cmat| -\log p(\gam)
\label{eq:ell2gam_general}
\end{align}
with $\Cmat \triangleq \lambda^{-1}\Imat+\Hmat\Gammat\Hmat^\hermit$ and $\Gammat = \diag(\gam)$. The \tii{} estimator of $\w$ follows as
\begin{align}
\what_{II}(\y) = \langle \w \rangle_{p(\w|\y;\gamhat_{II}(\y))} = \big(\Hmat^\hermit\Hmat+ \lambda^{-1}\Gammathat_{II}^{-1}\big)^{-1} \Hmat^\hermit\y,
\label{eq:type2w_general}
\end{align}
where $\Gammathat_{II}=\diag(\gamhat_{II}(\y))$ and $\langle \cdot \rangle_{p(\vect{x})}$ denotes expectation over the pdf $p(\vect{x})$. The impact of $p(\gam)$ on the estimator $\what_{II}$ is not straightforward. This complicates the task of selecting $p(\gam)$ inducing a sparse estimate of $\w$. In \cite{Wipf2011}, the relationship between \ti{} and \tii{} estimation has been identified. This result makes it possible to compare the two estimation methods. Invoking \cite[Theorem 2]{Wipf2011}, $\what_{II}(\y)$ is equivalently the minimizer of the \tii{} cost function
\begin{align}
\mathcal{L}_{II}(\w) \triangleq  \rho\|\y-\Hmat\w\|^2_2+\lambda^{-1}q_{II}(\w)
\label{eq:type2cost_general}
\end{align}
with penalty
\begin{align}
q_{II}(\w) 
=\min_{\gam} \big\{ \rho\w^\hermit\Gammat^{-1}\w+\rho\log|\Cmat|-\log p(\gam) \big\}.
\label{eq:qii_a_general}
\end{align} 
Specifically, $\what_{II}(\y)$ in \eqref{eq:type2w_general} equals the global minimizer of $\mathcal{L}_{II}(\w)$ iff $\gamhat_{II}(\y)$ equals the global minimizer of $\mathcal{L}_{II}(\gam)$. Likewise,  $\what_{\star}(\y) = \langle \w \rangle_{p(\w|\y;\gamhat_{\star}(\y))}$ is a local minimizer of $\mathcal{L}_{II}(\w)$ iff $\gamhat_{\star}(\y)$ is a local minimizer of $\mathcal{L}_{II}(\gam)$. 

The MAP estimates in \eqref{eq:wmap} and \eqref{eq:type2gam} cannot usually be computed in closed-form and one must resort to iterative inference methods to approximate them. One method is the Relevance Vector Machine (RVM) \cite{Tipping2001,WipfRao04}. In RVM the mixing pdf $p(\gamma_i)$ is equal to an improper constant prior.
An instance of the expectation-maximization (EM) algorithm is then formulated to approximate the \tii{} estimator. 
Another method, devised for real signal models in \cite{Figueiredo}, uses the EM algorithm to approximate two popular \ti{} estimators with respectively $\ell_1$-norm and log-sum constrained penalization. These penalization terms arise from selecting the mixing pdf equal to an exponential pdf and the noninformative Jeffreys prior, respectively. In the former case, the marginal prior pdf $p(\w)$ is the product of Laplace pdfs and $\mathcal{L}_{I}(\w)$ equals the cost function of Least Absolute Shrinkage and Selection Operator (LASSO) \cite{Tibshirani1994} or Basis Pursuit Denoising \cite{Chen1998}.\footnote{Let us point out that the hierarchical representation resulting in the $\ell_1$-norm presented in \cite{Figueiredo} is only valid for real-valued variables. In this paper, we extend this representation to cover complex-valued variables as well.}  

The sparse estimators in \cite{Tipping2001,WipfRao04,Figueiredo} inherit the limitation of the instances of the EM algorithm that they embed: high computational complexity and slow convergence \cite{Tipping2003}. 
To circumvent this shortcoming, a fast inference framework is proposed in \cite{Tipping2003} for RVM and later applied to derive the Fast Laplace algorithm \cite{Babacan2010}. The latter algorithm is derived based on an augmented probabilistic model obtained by adding a third layer to the real GSM model of the Laplace pdf; the third layer introduces a hyper-hyperprior for the rate parameter of the exponential pdf, which coincides with the regularization parameter of the $\ell_1$ penalization induced by the Laplace prior. However, as Fast Laplace is based on \tii{} estimation it cannot be seen as the adaptive Bayesian version of the $\ell_1$ re-weighted LASSO algorithm \cite{Candes2008}. The Bayesian version of this latter estimator is proposed in \cite{Griffin2007,Lee2010}. 

Even though the fast algorithms in \cite{Tipping2003} and \cite{Babacan2010} converge faster than their EM counterparts, they still suffer from slow convergence, especially in low and moderate signal-to-noise ratio (SNR) regimes as we will demonstrate in this paper. Furthermore, in these regimes the algorithms significantly overestimate the number of nonzero weights. We will show that this behavior is, in fact, a consequence of the prior models used to derive the algorithms.  

Coming back to the original motivation of this work, though complex GSM models have been proposed in the literature \cite{Palmer2009,Rakvongthai2010}, they have not been extensively applied within the framework of SBL. An example illustrating this fact is the hierarchical modeling of the $\ell_1$-norm in \ti{} estimation. While this penalty results from selecting the exponential mixing pdf for the entries in $\gam$ in real GSM models, said pdf will not induce the $\ell_1$-norm penalty for complex models. Yet to the best of our knowledge, the complex GSM model realizing the $\ell_1$-norm penalty has not been established in the literature. Moreover, it is not evident what sparsity-inducing property the complex GSM model exhibits when applied in \tii{} estimation. Motivated by the relative scarcity of formal tools for sparse learning in complex models and inspired by the recent analysis of sparse Bayesian algorithms in \cite{Wipf2011}, we propose and investigate an SBL approach that applies to both real and complex signal models. 

Starting in Section~\ref{sec:prior_models}, we first present a GSM model for both real and complex sparse signal representation where the mixing pdf $p(\gamma_i)$ is selected to be a gamma pdf. When $\w$ is real, the marginal prior pdf $p(\w)$ equals the product of Bessel K pdfs \cite{Barndorff-Nielsen1982,Gneiting1997,Eltoft2006}.\footnote{The Bessel K pdf is in turn a special case of even a larger class of generalized hyperbolic distributions \cite{Barndorff-Nielsen1982}, obtained when the mixing pdf is a Generalized Inverse Gaussian pdf.} We extend the Bessel K model to cover complex weights and model for this extension several penalty functions previously introduced for inferring real sparse weights. One important example is the hierarchical prior modeling inducing the $\ell_1$-norm penalty for complex weights. 
We then analyze the \ti{} and \tii{} estimators derived from the Bessel K model. We show that a sparsity-inducing prior for \ti{} estimation does not necessarily have this property for \tii{} estimation and, interestingly, a sparsity-inducing prior for real weights is not necessarily sparsity-inducing for complex weights. In the particular case where the dictionary matrix $\Hmat$ is orthonormal, we demonstrate, using the EM algorithm, that \ti{} and \tii{} estimators derived using the Bessel K model are generalizations of the soft-thresholding rule with degree of sparseness depending on the selection of the shape parameter of the gamma pdf $p(\gamma_i)$. Additionally, we show that this model has a strong connection to the Bayesian formalism of the group LASSO \cite{Lee2010,Kyung2010}. Note that the Bessel K model has been previously introduced for sparse signal representation \cite{Caron2008,Griffin2010}. However, these works were restricted to the inference of real weights and did not consider the relationship between \ti{} and \tii{} estimation.

In Section~\ref{sec:sbl}, we propose greedy, low-complexity estimators using the Bessel K model. The estimators are based on a modification of the EM algorithm for \tii{} estimation.   
As the Bessel K model encompasses the prior models used in \cite{Tipping2003} and \cite{Babacan2010}, the iterative algorithms derived in these publications can be seen as instances of our estimators for particular settings of the associated parameters of the gamma mixing pdf.  

Section~\ref{sec:simulations} provides numerical results obtained via Monte Carlo simulations that reveal the superior performance of the proposed estimators in terms of convergence speed of the algorithms, sparseness, and mean-squared error (MSE) of the estimates. Furthermore, and of great importance in many engineering areas, the estimators show a significant robustness in low and moderate SNR regimes; a property not exhibited by the traditional SBL estimators, like \cite{Tipping2003} and \cite{Babacan2010}, and other state-of-the-art non-Bayesian sparse estimators. This result opens for a potential application of our estimators in systems operating in these SNR regimes - e.g., receivers in wireless communications \cite{Pedersen2012,Pedersen2013}. Furthermore, the proposed estimators can inherently incorporate the estimation of the noise variance. In the literature this parameter is often learned from a training procedure or tuned for optimality. Since the algorithms in \cite{Tipping2003} and \cite{Babacan2010} only differ from ours in the choice of parameters of the mixing pdf, we can safely conclude that the observed performance benefits are a direct consequence of our proposed prior model. 

Finally, we conclude the paper in Section~\ref{sec:conclusion}.
 
\section{The Bessel K Model for Real and Complex Signal Representation \label{sec:prior_models}}
In this section we present the Bessel K model for SBL. We first state the probabilistic model of the signal model \eqref{eq:model}. Based on this probabilistic model we analyze the \ti{} and \tii{} cost functions. We then show how to obtain various estimators with different sparsity-inducing properties by appropriately setting the parameters of the Bessel K model. 

\subsection{Probabilistic Model}

We begin with the specification of the probabilistic model for \eqref{eq:model} augmented with the two-layer prior model for $\w$:
\begin{align}
p(\y,\w,\gam) = p(\y|\w)p(\w|\gam)p(\gam).
\label{eq:jointpdf2layer}
\end{align}
From \eqref{eq:model}, $p(\y|\w) = \GaussPDF(\y|\Hmat\w,\lambda^{-1}\matr{I})$ if the signal model is real and $p(\y|\w) = \CGaussPDF(\y|\Hmat\w,\lambda^{-1}\matr{I})$ if the model is complex.\footnote{$\GaussPDF(\cdot|\vect{a},\matr{B})$ and $\CGaussPDF(\cdot|\vect{a},\matr{B})$ denote respectively a multivariate real and a multivariate complex Gaussian pdf with mean vector $\vect{a}$ and covariance matrix $\matr{B}$. We shall also make use of the gamma pdf $\GammaPDF(\cdot|a,b)= \frac{b^a}{\Gamma(a)}x^{a-1}\exp(-bx)$ with shape parameter $a$ and rate parameter $b$.} 

The sparsity constraints on $\w$ are determined by the joint prior pdf $p(\w|\gam)p(\gam)$. Motivated by previous works on GSM modeling and SBL \cite{Tipping2001,WipfRao04,Figueiredo} we select the conditional prior pdf $p(\w|\gam)$ to factorize in a product of zero-mean Gaussian pdfs: $p(\w|\gam) = \prod_i p(w_i|\gamma_i)$ where
\begin{align}
p(w_i|\gamma_i)=\big(\frac{\rho}{\pi\gamma_i}\big)^\rho\exp\big(-\rho\frac{|w_i|^2}{\gamma_i}\big). 
\label{eq:real_complex_gaussian}
\end{align} 
In the above expression, $\rho=1/2$ when $\w$ is real and $\rho=1$ when $\w$ is complex. We choose the mixing pdf $p(\gam)$ to be a product of identical gamma pdfs, i.e., $p(\gam)=\prod_i p(\gamma_i;\epsilon,\eta)$ with $p(\gamma_i;\epsilon,\eta) \triangleq \GammaPDF(\gamma_i|\epsilon,\eta)$. The prior pdf for $\w$ is then given by $p(\w;\epsilon,\eta)=\int p(\w|\gam)p(\gam;\epsilon,\eta)\mathrm{d}\gam=\prod_i p(w_i;\epsilon,\eta)$ 
with  
\begin{align}
p(w_i;\epsilon,\eta)= \frac{2(\rho\eta)^{\frac{(\epsilon+\rho)}{2}}}{\pi^\rho \Gamma(\epsilon)}|w_i|^{\epsilon -\rho}K_{\epsilon-\rho}(2\sqrt{\rho\eta}|w_i|).
\label{eq:palpha}
\end{align}
In this expression, $K_\nu(\cdot)$ is the modified Bessel function of the second kind and order $\nu \in \realnum$. In case $\w$ is real ($\rho = 1/2$), we obtain the GSM model of the Bessel K pdf \cite{Barndorff-Nielsen1982,Gneiting1997}. We will keep the same terminology when $\w$ is complex ($\rho=1$).\footnote{To the authors' best knowledge, the GSM model of the Bessel K pdf has only been presented for real variables.} The Bessel K pdf \eqref{eq:palpha} represents a family of prior pdfs for $\w$ parametrized by $\epsilon$ and $\eta$. By selecting different values for $\epsilon$ and $\eta$, we realize various penalty functions for \ti{} and \tii{} estimation as shown in the following.

\subsection{\ti{} Cost Function}

The \ti{} cost function $\mathcal{L}_{I}(\w)$ induced by the Bessel K model is given by \eqref{eq:type1cost} with penalty $q_I(\w) = \sum_i q_I(w_i;\epsilon,\eta)$ where
\begin{align}
q_I(w_i;\epsilon,\eta) \triangleq -\log\left(|w_i|^{\epsilon-\rho} K_{\epsilon-\rho}\left(2\sqrt{\rho\eta}|w_i|\right)\right).
\label{eq:qi}
\end{align} 
Special cases of \ti{} penalties resulting from the Bessel K pdf have already been considered in the literature for sparse regression when the weights are real \cite{Caron2008,Griffin2010}. We review them together with introducing the corresponding extension to complex weights.
    
\subsubsection{The $\ell_1$-norm penalty}

This penalty is of particular importance in sparse signal representation as the convex relaxation of the $\ell_0$-norm.\footnote{The $\ell_0$-norm of the vector $\vect{x}$ is the number of nonzero entries in $\vect{x}$. Note that by abuse of terminology $\|\cdot\|_0$ is coined a norm even though it does not fulfill all properties of a norm.}

When $\w$ is real, it is well-known that the Laplace prior induces the $\ell_1$-norm penalty. The Bessel K pdf \eqref{eq:palpha} encompasses the Laplace pdf as a special case with the selection $\epsilon=1$ and $\rho=1/2$:\footnote{Here, we make use of the identity $K_{\frac{1}{2}}(z) = \sqrt{\frac{\pi}{2z}}\exp(-z)$ \cite{Abramowitz}.}
\begin{align}
p(w_i;\epsilon=1,\eta) = \sqrt{\frac{\eta}{2}}\exp(-\sqrt{2\eta}|w_i|),\quad w_i \in \mathbb{R}.
\label{eq:laplace_real} 
\end{align}
The Laplace pdf for real weights is thereby the pdf of a GSM with an exponential mixing pdf \cite{Andrews1974}. 

The extension of \eqref{eq:laplace_real} to $\w$ complex is not straightforward. One approach is to treat the real and imaginary parts of each $w_i$ independently with both parts modeled according to the real GSM representation of the Laplace pdf. Doing so using \eqref{eq:laplace_real} we obtain $p(w_i) = \frac{\eta}{2}\exp(-\sqrt{2\eta}(|\Re\{w_i\}|+|\Im\{w_i\}|))$. Obviously this approach does not lead to the $\ell_1$-norm penalty for \ti{} estimation.\footnote{The $\ell_1$-norm for the complex vector $\vect{x}$ is defined as $\|\vect{x}\|_1 = \sum_i|x_i| = \sum_i\sqrt{\Re^2\{x_i\}+\Im^2\{x_i\}}$ \cite{Kim2007,Wright2009}.} The complex GSM model with a gamma mixing pdf with shape parameter $\epsilon=3/2$ does induce this penalty. Indeed, with this setting, \eqref{eq:palpha} becomes for $\rho=1$
\begin{align}
p(w_i;\epsilon=3/2,\eta) = \frac{2\eta}{\pi}\exp(-2\sqrt{\eta}|w_i|),\quad w_i \in \mathbb{C}.
\label{eq:laplace_complex}
\end{align}
Throughout the paper, we refer to the pdf in \eqref{eq:laplace_complex} as the Laplace pdf for complex weights.

In summary, the Bessel K model induces the $\ell_1$-norm penalty $q_I(\w) = 2\sqrt{\rho\eta} \sum_i |w_i|$ with the selection $\epsilon=\rho+1/2$. The introduced GSM model of the Laplace pdf for both real and complex variables is strongly connected with the group LASSO and its Bayesian interpretation \cite{Lee2010,Kyung2010}, where sparsity is enforced simultaneously over groups of $k$ variables. In the Bayesian interpretation of the group LASSO a gamma pdf with shape parameter $(k+1)/2$ is employed to model the prior for each of the variables in a group. This choice of shape parameter is consistent with the choice of $\epsilon$ in the GSM model of the Laplace prior: in the real case a group consists of $k =1$ variable and, thus, $(k+1)/2 = 1$, whereas in the complex case, a group consists of the real and imaginary parts of a complex variable, hence, $k=2$ and $(k+1)/2 = 3/2$.  

\subsubsection{The log-sum penalty}

The selection $\epsilon=\eta=0$ in \eqref{eq:palpha} entails the Jeffreys (improper) prior density $p(\gamma_i)\propto \gamma^{-1}_i$ and thereby the improper marginal prior density  $p(\w)\propto \prod_i |w_i|^{-2\rho}$. Thus, when the mixing density of the GSM is chosen to be noninformative, the log-sum penalization $q_I(\w) = 2\rho\sum_i\log |w_i|$ is invoked in \eqref{eq:type1cost}. This penalty has gained much interest in the literature, including \cite{Tipping2001,WipfRao04,Figueiredo,Candes2008,Wipf2010}, as it is known to strongly promote sparse estimates.

\subsubsection{The Bessel K penalty} 

The Bessel K pdf can be used with arbitrary values of $\epsilon \geq 0$ controlling its sparsity-inducing property. 
To illustrate this, \Fig~\ref{fig:penaltyterms_type12}(a) depicts one contour line of the restriction\footnote{Let $f$ denote a function defined on a set $A$. The restriction of $f$ to a subset $B\subset A$ is the function defined on $B$ that coincides with $f$ on this subset.} to $\realnum^2$ of $q_{I}(w_1,w_2;\epsilon,\eta)$ in \eqref{eq:qi} for selected values of $\epsilon$. As $\epsilon$ approaches zero more probability mass concentrates along the $\w$-axes; as a consequence, the mode of the resulting posterior pdf $p(\w|\y;\epsilon,\eta)$ is more likely to be close to the axes, thus encouraging a sparse estimate. The behavior of the $\ell_1$-norm penalty that results from the selection $\epsilon=\rho+1/2 = 3/2$ is also clearly recognized.

\subsection{\tii{} Cost Function}

We invoke Theorem 2 in \cite{Wipf2011} to obtain the \tii{} cost function induced by the Bessel K model (see \eqref{eq:type2cost_general} and \eqref{eq:qii_a_general}):
\begin{align}
\mathcal{L}_{II}(\w) \triangleq  \rho\|\y-\Hmat\w\|^2_2+\lambda^{-1}q_{II}(\w)
\label{eq:type2cost}
\end{align}
with 
\begin{align}
q_{II}(\w;\epsilon,\eta) 
=\min_{\gam} \big\{ \rho\w^\hermit\Gammat^{-1}\w+\rho\log|\Cmat|+(1-\epsilon)\sum_i \log\gamma_i+\eta\sum_i\gamma_i \big\}.
\label{eq:qii_a}
\end{align} 

\begin{figure*}[!t]
\centering
\centerline{
\subfigure[\ti]{\includegraphics[width=0.35\linewidth]{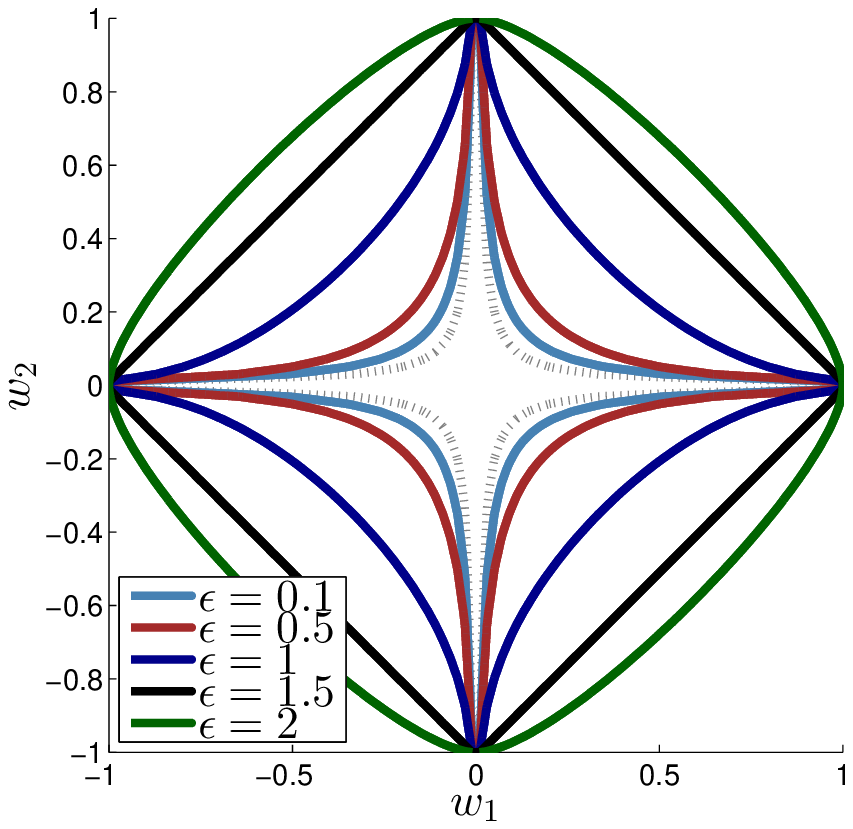}}
\hspace{1cm}
\subfigure[\tii]{\includegraphics[width=0.35\linewidth]{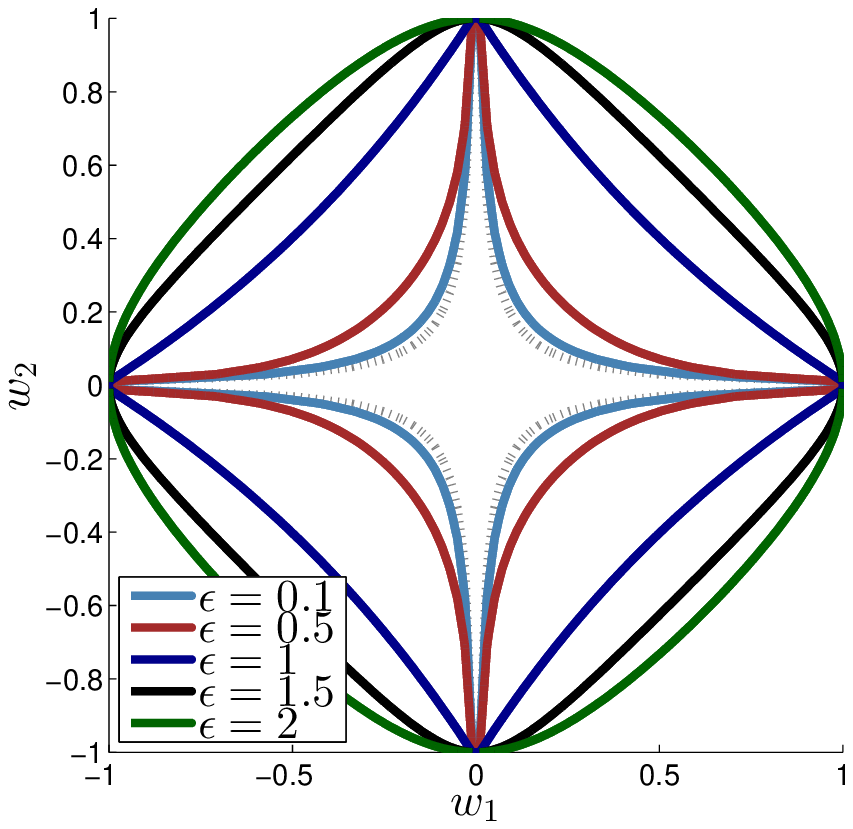}}
}
\caption{One contour line of the restriction to $\realnum^2$ of (a) $q_{I}(w_1,w_2;\epsilon,\eta)$ and (b) $q_{II}(w_1,w_2;\epsilon,\eta=1)$ for selected values of $\epsilon$. In (b) $\Hmat$ is orthonormal and $\lambda^{-1} = 1/4$. The gray dashed lines depict the contour lines corresponding to the setting $\epsilon=\eta=0$, i.e., the mixing density equals the Jeffreys prior.}
\label{fig:penaltyterms_type12}
\end{figure*}

In contrast to $q_I(\w)$, $q_{II}(\w)$ is nonseparable. This makes an interpretation of $q_{II}(\w)$ as done for $q_{I}(\w)$ in \Fig~\ref{fig:penaltyterms_type12}(a) rather difficult. However, this interpretation becomes straightforward if $\Hmat$ is orthonormal, i.e., $\Hmat^\hermit\Hmat = \matr{I}$. In this case $q_{II}(\w)$ is separable, i.e., $q_{II}(\w) = \sum_i q_{II}(w_i)$ with
\begin{align}
q_{II}(w_i;\epsilon,\eta) = \min_{\gamma_i} \big\{ \rho\frac{|w_i|^2}{\gamma_i}+\rho\log(\lambda^{-1}+\gamma_i)+(1-\epsilon)\log\gamma_i+\eta\gamma_i \big\}.  
\label{eq:qii_b}
\end{align}     
\Fig~\ref{fig:penaltyterms_type12}(b) shows the contours of the restriction to $\mathbb{R}^2$ of $q_{II}(w_1,w_2;\epsilon,\eta)$ in \eqref{eq:qii_b} for different values of $\epsilon$. Again, we observe the same increased concentration of mass around the $\w$-axes for decreasing values of $\epsilon$. Interestingly, $q_{II}(w_1,w_2;\epsilon=3/2,\eta)$ is no longer sparsity-inducing as compared to $q_{I}(w_1,w_2;\epsilon=3/2,\eta)$. \textit{Thus, a sparsity-inducing prior model for \ti{} estimation is not necessarily sparsity-inducing for \tii{} estimation.} We further investigate this important result in Section~\ref{sec:t1t2estimators}. 

\begin{figure*}[!t]
\centering
\centerline{
\subfigure[Bessel K, $\epsilon=1/2$]{\includegraphics[width=0.3\linewidth]{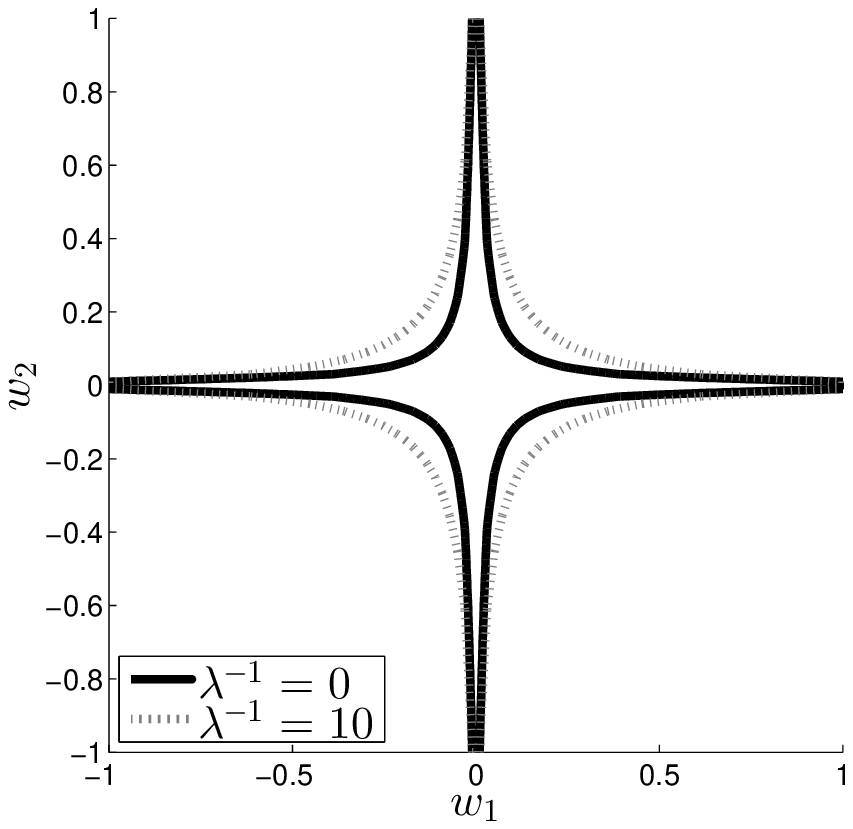}}
\hspace{0.25cm}
\subfigure[Bessel K, $\epsilon=1$]{\includegraphics[width=0.3\linewidth]{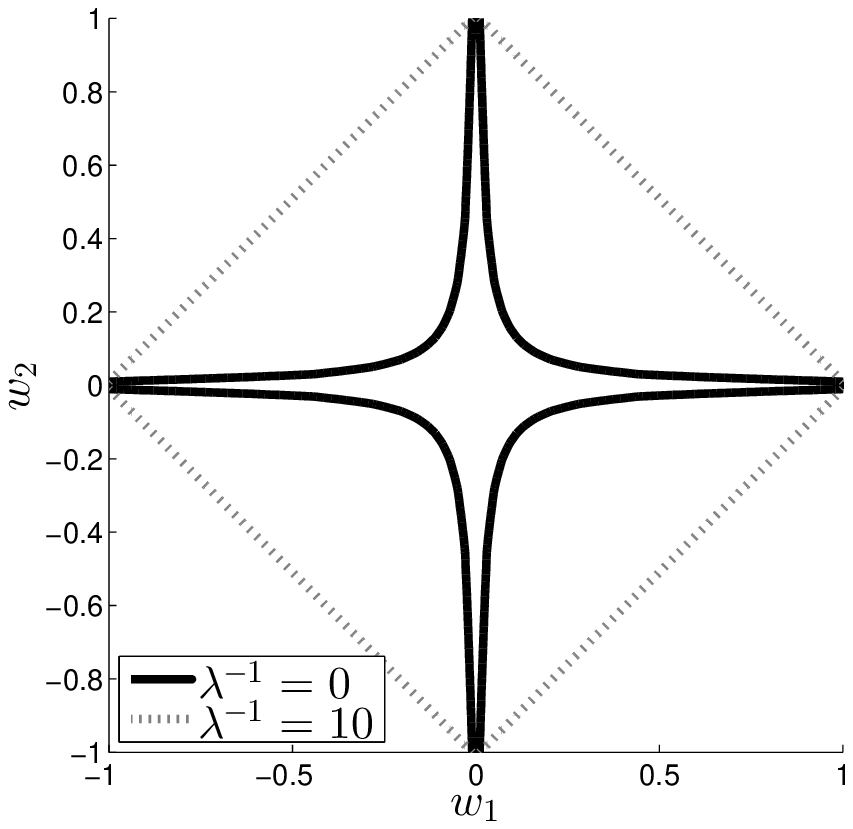}}
\hspace{0.25cm}
\subfigure[RVM]{\includegraphics[width=0.3\linewidth]{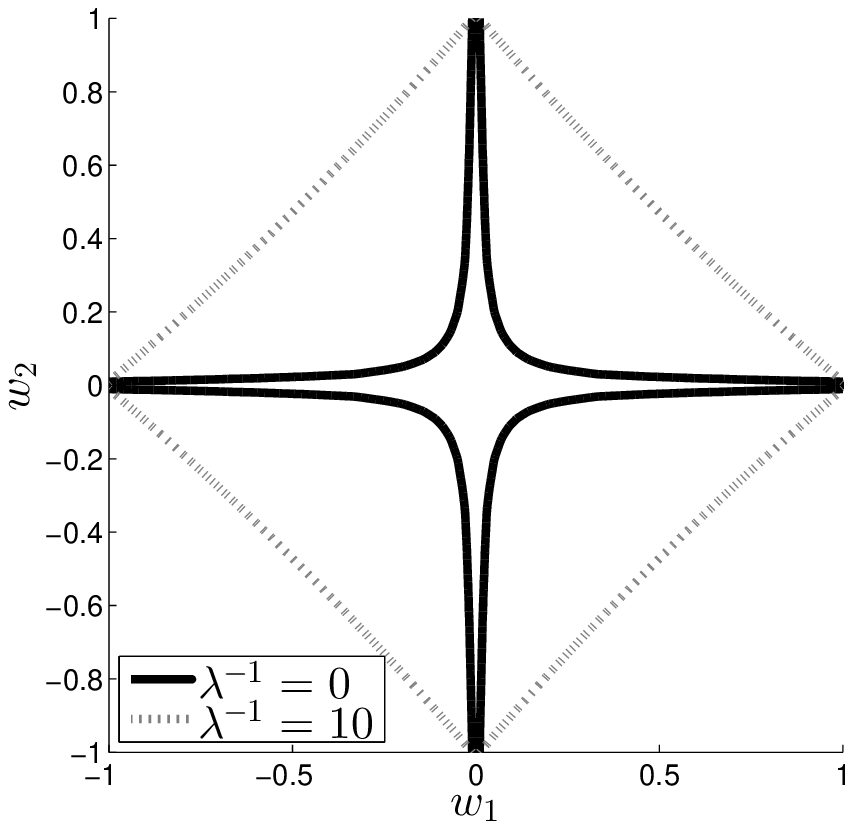}}
}
\caption{One contour line of the restriction to $\realnum^2$ of (a) $q_{II}(w_1,w_2;\epsilon=1/2,\eta=1)$, (b) $q_{II}(w_1,w_2;\epsilon=1,\eta=1)$, and (c) $q_{II}(w_1,w_2;\epsilon=1,\eta=0)$ with $\Hmat$ orthonormal and $\lambda^{-1}$ as a parameter. Note that $q_{II}(w_1,w_2;\epsilon=1,\eta=0)$ in (c) coincides with the penalty used in RVM \cite{Tipping2001,WipfRao04}.}
\label{fig:penaltyterms_type12sig}
\end{figure*} 

Another important property of the \tii{} penalty is its dependency on the noise variance $\lambda^{-1}$. 
\Fig~\ref{fig:penaltyterms_type12sig}(a) and \Fig~\ref{fig:penaltyterms_type12sig}(b) depict a single contour line of  \eqref{eq:qii_b} for two values of $\epsilon$ and two values of $\lambda^{-1}$.  
Notice that $q_{II}(\w;\epsilon=1/2,\eta=1)$ resembles the log-sum penalty even in noisy conditions. For comparison purposes, we show in \Fig~\ref{fig:penaltyterms_type12sig}(c) the \tii{} penalty computed with the prior model in RVM \cite{Tipping2001,WipfRao04} which utilizes a constant prior pdf $p(\gamma_i)\propto 1$ (corresponding to setting $\epsilon=1$ and $\eta=0$ in \eqref{eq:qii_a}). When $\lambda^{-1}=0$ the RVM penalty equals the log-sum penalty. However, in noisy conditions the RVM penalty resembles the $\ell_1$-norm penalty. Note that we cannot simply set $\lambda^{-1}$ to some small value in order to obtain a strong sparsity-inducing penalty in RVM as $\lambda^{-1}$ acts as a regularization of $q_{II}(\w)$ in \eqref{eq:type2cost}. Based on this observation, we expect that the \tii{} estimator derived from the Bessel K model achieves improved sparsity performance as compared to RVM in noisy scenarios. The numerical results conducted in Section~\ref{sec:simulations} demonstrate that this is indeed the case.

\subsection{\ti{} and \tii{} Estimation \label{sec:t1t2estimators}}

Having evaluated the impact of $\epsilon$ on $q_{I}(\w)$ and $q_{II}(\w)$, we now investigate its effect of this parameter on the corresponding \ti{} and \tii{} estimators. We demonstrated that as $\epsilon$ decreases, $q_{I}(\w)$ and $q_{II}(\w)$ become more and more sparsity-inducing which motivates the selection of a small $\epsilon$ for sparse estimation. On the other hand the Bessel K model for \ti{} and \tii{} estimation dominates the information contained in the observation $\y$ for decreasing values of $\epsilon$. Specifically, in case of \ti{}, when $\epsilon\leq \rho$ then $\lim_{w_i\rightarrow 0} q_I(w_i) = -\infty$, hence, the \ti{} estimator does not exist as $\mathcal{L}_I(\w)$ has singularities. Likewise, this is the case for the \tii{} estimator when $\epsilon<1$. 
The unbounded behavior of these penalties naturally questions the practicability of the Bessel K model in SBL. At least one would expect that we should refrain from selecting $\epsilon\leq\rho$ in case of \ti{} estimation and $\epsilon<1$ for \tii{} estimation. Note, however, that utilizing unbounded penalties in SBL is not uncommon. Examples include \cite{Caron2008,Griffin2010} as well as the popular GSM model realizing the log-sum penalty in e.g., \cite{Figueiredo}. Furthermore, the sparsity-inducing behavior of the penalty curves in \Fig~\ref{fig:penaltyterms_type12} and \Fig~\ref{fig:penaltyterms_type12sig} provides a strong motivation for using the Bessel K model in SBL. The approach is to formulate approximate inference algorithms, such as EM, for \ti{} and \tii{} estimation that overcome the difficulty of the singularities in the objective functions.

\subsubsection{Approximate \ti{} estimation}

The EM algorithm approximating the \ti{} estimator makes use of the complete data $\{\gam,\y\}$ for $\w$.\footnote{This EM algorithm is derived in \ref{app:sbl_em1}.} The M-step computes an estimate of $\w$ as the maximizer of
\begin{align}
 \langle \log p(\y|\w)p(\w|\gam)p(\gam)\rangle_{p(\gam;\what)},
\label{eq:emcost1}
\end{align}
where $p(\gam;\what)$ is computed in the E-step. Notice that as $p(\w|\y,\gam)\propto p(\y|\w)p(\w|\gam)$ is proportional to a Gaussian pdf for $\w$, \eqref{eq:emcost1} does not have any singularity in contrast to $\mathcal{L}_I(\w)$.

\begin{figure*}[!t]
\centering
\centerline{
\subfigure[\ti]{\includegraphics[width=0.45\linewidth]{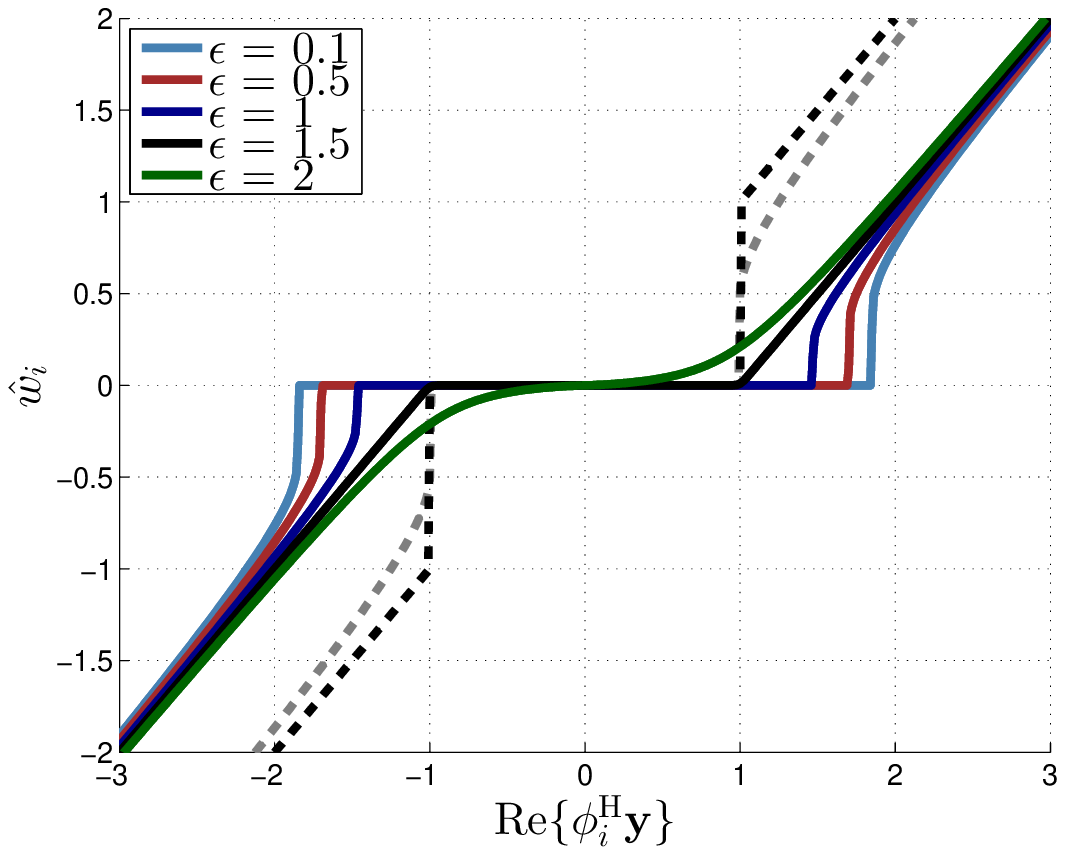}}
\hspace{0.5cm}
\subfigure[\tii]{\includegraphics[width=0.45\linewidth]{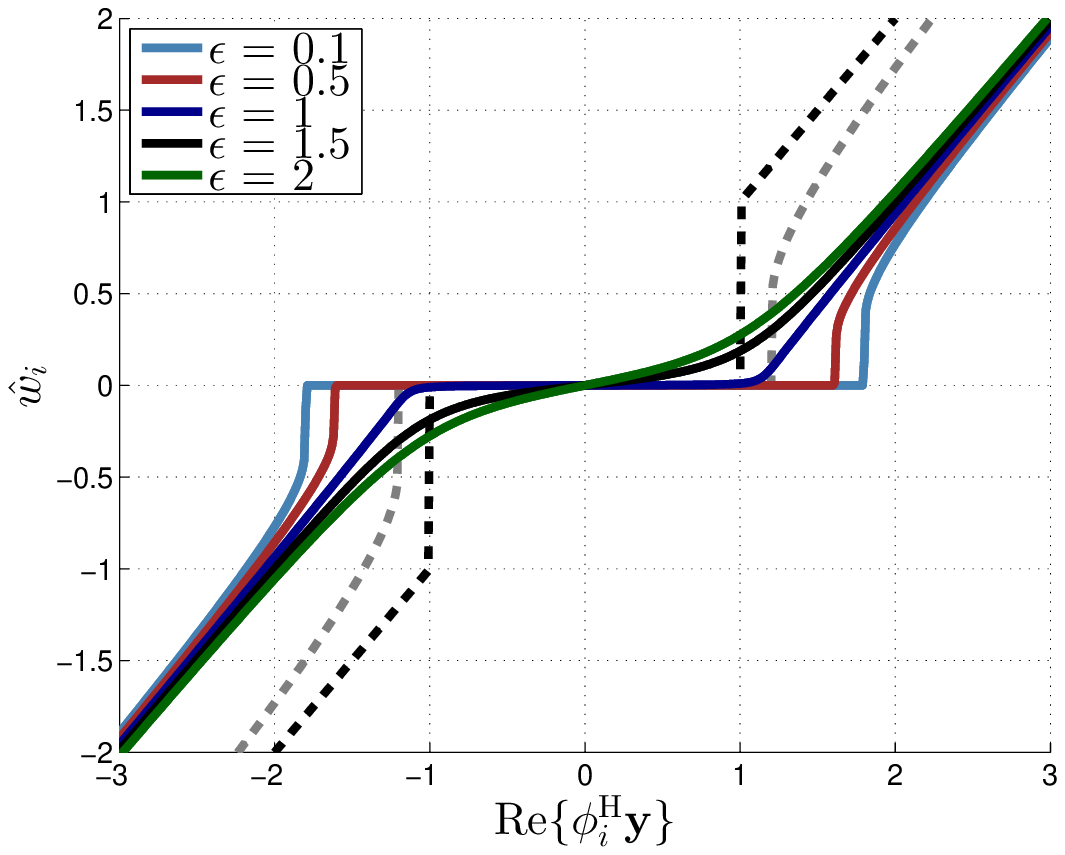}}
}
\caption{Restriction to $\Im\{\h_i^\hermit\y\}=0$ of the EM-based \ti{} and \tii{} estimators of the complex weight $w_i$ when $\Hmat$ is orthonormal. The gray dashed lines depict the estimator corresponding to the setting $\epsilon=\eta=0$, i.e., when $p(\gamma_i)$ equals the Jeffreys prior. The black dashed lines represent the hard-threshold rule. All results were generated using $\lambda^{-1} = 1/4$ and $\eta$ set such that $\lambda^{-1}\sqrt{\eta/\rho} = 1$.}
\label{fig:estimationrules}
\end{figure*}

In order to get further insight into the impact of $\epsilon$ on the EM algorithm, we follow \cite{Figueiredo} and let $\Hmat$ be orthonormal such that the EM update of the estimate of $\w$ decouples into $N$ independent scalar optimization problems. \Fig~\ref{fig:estimationrules}(a) visualizes the EM estimator for different values of $\epsilon$. Clearly, the EM estimator approximates the soft-thresholding rule for large values of $\Re\{\h_i^\hermit\y\}$ and as $\epsilon$ decreases the threshold value increases, thus, encouraging sparsity.

When the Bessel K pdf equals the Laplace pdf (i.e., $\epsilon = \rho + 1/2$), $\what_{I}$ coincides with the soft-thresholding rule, which can be computed in closed form: 
\begin{align}
\hat{w}_{I,i}(\y) = 
\mathrm{sign}(\h_i^\hermit\y)\mathrm{max}\left\{0, |\h_i^\hermit\y| - \lambda^{-1}\sqrt{\frac{\eta}{\rho}}\right\}, \quad i=1,\ldots,N.
\label{eq:softrule}
\end{align}  
Here, $\mathrm{sign}(x)=x/|x|$ is the sign function. Notice that the EM estimator with $\epsilon= \rho+1/2$ approximates \eqref{eq:softrule} as depicted in \Fig~\ref{fig:estimationrules}(a).

\subsubsection{Approximate \tii{} estimation}

The EM algorithm approximating \tii{} estimation is devised using $\{\w,\y\}$ as the complete data for $\gam$.\footnote{This EM algorithm is derived in Section~\ref{sec:sbl_em2}.} The M-step computes an estimate of $\gam$ as the maximizer of
\begin{align}
 \langle \log p(\y|\w)p(\w|\gam)p(\gam)\rangle_{p(\w;\gamhat)},
\label{eq:emcost2}
\end{align}
with $p(\w|\gamhat)$ computed in the E-step. As $p(\gam|\w)\propto p(\w|\gam)p(\gam)$ is a Generalized Inverse Gaussian (GIG) pdf for $\gam$, \eqref{eq:emcost2} does not exhibit any singularity as opposed to $\mathcal{L}_{II}(\gam)$.

In \Fig~\ref{fig:estimationrules}(b), we show the EM estimate of $w_i$ for different settings of $\epsilon$. Similar to \ti, the \tii{} estimate approaches the soft-thresholding rule as $\Re\{\h_i^\hermit\y\}$ becomes larger and as $\epsilon$ decreases a sparser estimate is obtained. However, when $\epsilon = 3/2$, i.e., utilizing the Laplace GSM model for the complex weights, the \ti{} estimator coincides with the soft-threshold rule while the \tii{} estimator does not have this threshold-like behavior and is not sparse. This was already indicated by the behavior of $q_{II}(\w;\epsilon=3/2,\eta)$ in \Fig~\ref{fig:penaltyterms_type12}(b).

\begin{figure*}[!t]
\centering
\centerline{
\subfigure[$\rho = 1/2$, $\epsilon = 1$]{\includegraphics[width=0.23\linewidth]{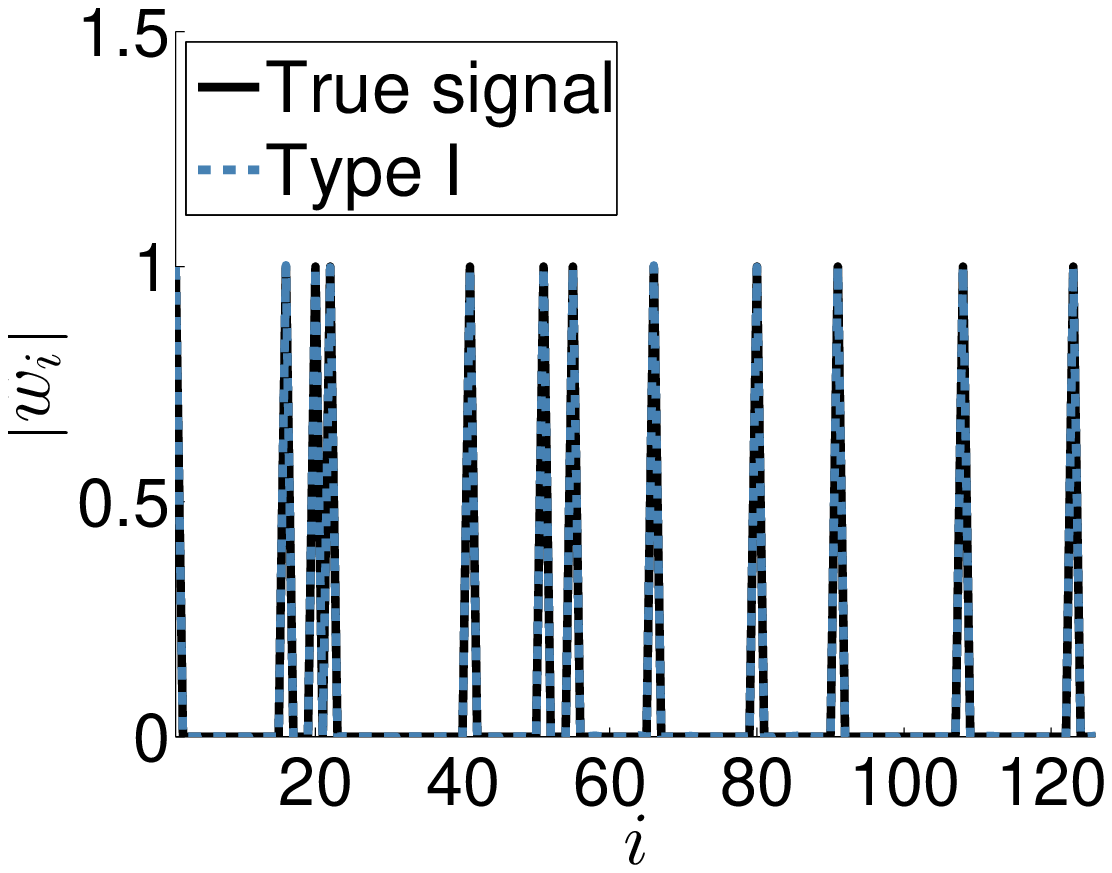}}
\subfigure[$\rho = 1/2$, $\epsilon = 1$]{\includegraphics[width=0.23\linewidth]{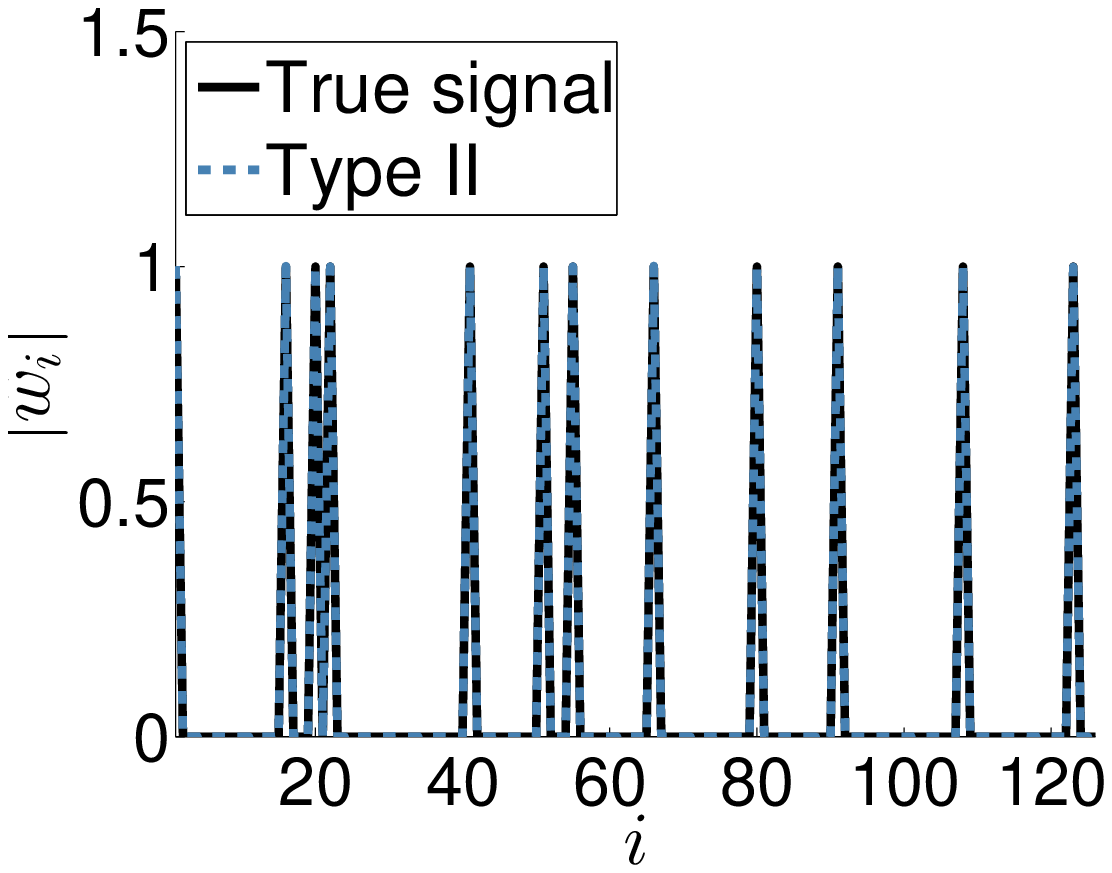}}
\subfigure[$\rho = 1$, $\epsilon = 3/2$]{\includegraphics[width=0.23\linewidth]{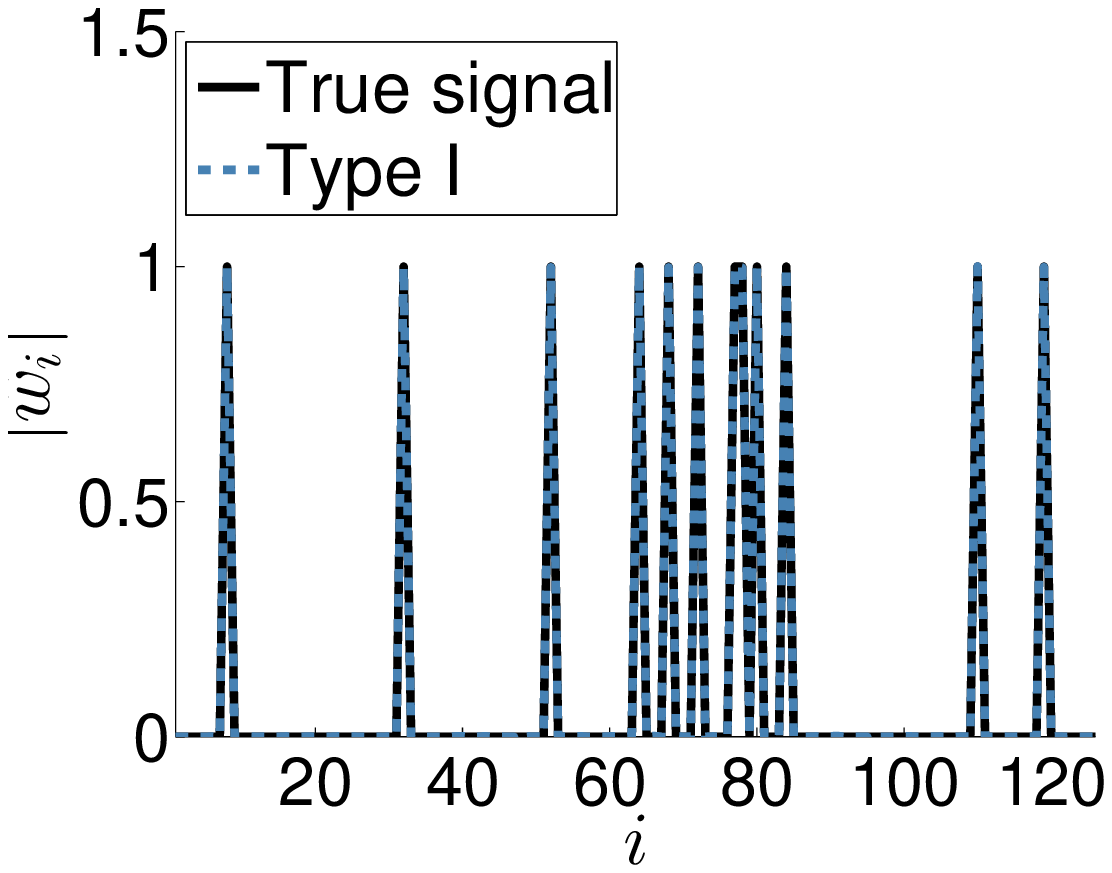}}
\subfigure[$\rho = 1$, $\epsilon = 3/2$]{\includegraphics[width=0.23\linewidth]{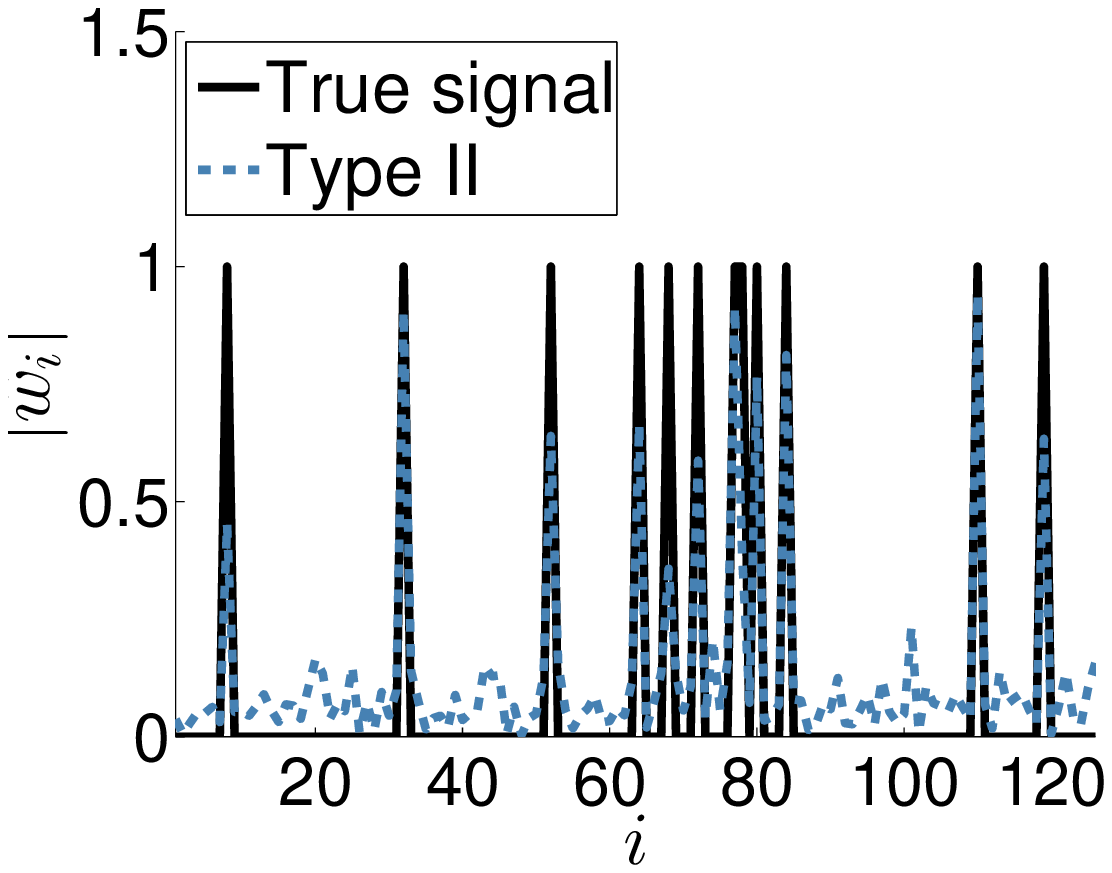}}
}
\caption{EM-based \ti{} and \tii{} estimates using the Laplace GSM model for (a)-(b) real and (c)-(d) complex weights. For these simulations, $\Hmat\in\compnum^{50\times128}$ with its entries drawn independently according to $\phi_{mi}\sim \CGaussPDF(0,1/M)$. The $K=12$ nonzero entries in $\w$ are of the form $w_k=\exp(j\theta_k)$ with $\theta_k$, $k=1,\ldots,K$, drawn independently according to a uniform distribution on $[0,2\pi)$. The SNR is fixed at 60 dB.}
\label{fig:simlaplace}
\end{figure*}

From \Fig~\ref{fig:estimationrules} we conclude that the EM-based \ti{} estimator is a sparse estimator for $\epsilon \leq \rho + 1/2$, whereas the EM-based \tii{} estimator only exhibits this property for $\epsilon \leq 1$. In \Fig~\ref{fig:simlaplace}, we illustrate this important difference in the behavior of these estimators for real and complex signal representation when utilizing the GSM model of the Laplace prior: the EM-based \ti{} estimator achieves  a sparse solution for both real and complex weights, whereas for the EM-based \tii{} estimator this is only the case for real weights.

\section{Sparse Bayesian Inference \label{sec:sbl}}
In this section we derive a Bayesian inference scheme that relies on the Bessel K model presented in Section~\ref{sec:prior_models}. 
First, we obtain an EM algorithm that approximates the \tii{} estimator of the weight vector $\w$ in \eqref{eq:type2w_general}. Inspired by \cite{Tipping2003} and \cite{ShutinFastRVM} we then derive a fast algorithm based on a modification of the EM algorithm. We show that this algorithm actually encompasses the fast algorithms in \cite{Tipping2003} and \cite{Babacan2010} as special instances. 

Naturally, the approach presented here can also be applied to derive algorithms approximating the \ti{} estimator. However, numerical investigations not reported here indicate that these algorithms often fail to produce sparse estimates of $\w$ when small values of the parameter $\epsilon$ are selected. Hence, we restrict the discussion in this section to algorithms approximating the \tii{} estimator.

\subsection{Sparse Bayesian Inference Using EM \label{sec:sbl_em2}}

We adapt the EM algorithm approximating the \tii{} estimator previously used for SBL \cite{Tipping2001,Tipping2003,WipfRao04,Ji2008,Babacan2010} to the Bessel K model. As the value of $\lambda$ is in general unknown and has a significant impact on the sparsity-inducing property on $q_{II}(\w)$ (see Section~\ref{sec:prior_models}), we include the estimation of this parameter in the inference framework. We seek the MAP estimate of $\{\gam,\lambda\}$, i.e., the maximizer of 
\begin{align}
	\mathcal{L}(\gam,\lambda) 
	= \log p(\y,\gam,\lambda)
= \log(p(\y|\gam,\lambda)p(\gam)p(\lambda)).	
	\label{eq:sbl_cost}
\end{align}
We use the EM algorithm to approximate the MAP estimator. We specify $\{\w,\y\}$ to be the complete data for $\{\gam,\lambda\}$. With this choice the E-step of the EM algorithm computes the conditional expectation
\begin{align}
  \langle \log p(\y,\w,\gam,\lambda)\rangle_{p(\w|\y,\gam^{[t]},\lambda^{[t]}) }
 \label{eq:completeloglikelihood}
\end{align}
with $p(\w|\y,\gam^{[t]},\lambda^{[t]}) = \GaussPDF( \w|\muv^{[t]}, \Sigmamat^{[t]} )$  or $p(\w|\y,\gam^{[t]},\lambda^{[t]}) = \CGaussPDF( \w|\muv^{[t]}, \Sigmamat^{[t]} )$ depending on whether the underlying signal model is real or complex. Here, $(\cdot)^{[t]}$ denotes the estimate of the parameter given as an argument at iteration $t$.  
In either case, the parameters of the conditional pdf of $\w$ read
\begin{align}
	\Sigmamat^{[t]} &= \big(\lambda^{[t]} \Hmat^\hermit\Hmat+ (\Gammat^{[t]})^{-1}\big)^{-1}, \label{eq:wcov}\\
	\muv^{[t]} &= \lambda^{[t]} \Sigmamat^{[t]}\Hmat^\hermit\y.  \label{eq:wmean}
\end{align}
The M-step of the EM algorithm updates the estimate of $\{\gam,\lambda\}$ as the maximizer of \eqref{eq:completeloglikelihood}: 
\begin{align}
	\gamma^{[t+1]}_i &=  \frac{\epsilon-\rho-1 + \sqrt{(\epsilon-\rho-1)^2+4\rho\eta\langle|w_i|^2\rangle^{[t]}}}{2\eta}, &\quad i = 1,\ldots,N, \label{eq:gamma_mode}\\
	\lambda^{[t+1]} &=  \frac{M}{\|\y-\Hmat\muv^{[t]}\|^2_2+\mathrm{tr}(\Hmat^\hermit\Hmat\Sigmamat^{[t]})} \label{eq:lambda_mode}.
\end{align}
Here, $\langle|w_i|^2\rangle^{[t]}$ is the $i$th diagonal element of $\Sigmamat^{[t]} + \muv^{[t]}(\muv^{[t]})^\hermit$ computed in the E-step and $\mathrm{tr}(\cdot)$ is the trace operator.
 
\subsection{Modified update of $\gamma^{[t+1]}_i$ \label{subsec:fastscheme}}
One of the major drawbacks of the EM algorithm approximating the \tii{} estimator is its slow convergence, as observed in, e.g., \cite{Tipping2003}.\footnote{The selected mixing pdf also has a significant impact on the convergence speed as shown in Section~\ref{sec:simulations}.} In this section, we discuss a modification of the EM algorithm that improves the convergence speed. The proposed algorithm is inspired by \cite{Tipping2003} and \cite{ShutinFastRVM}. 
To this end, we focus on the update of a single estimate of $\gamma_i$ and express this update as a (non-linear recurrent) function of the previous update. Then, we analyze the fixed points of this function for different settings of the hyperparameters $\epsilon$ and $\eta$ and formulate a new update rule for the estimate of $\gamma_i$ at iteration $t~+~1$ based on these fixed points.  
From this point on, we restrict our analysis to the Bessel K model with $\epsilon\leq 1$ since, as discussed in Section~\ref{sec:prior_models}, the setting $\epsilon > 1$ does not yield a sparse \tii{} estimator.

To begin, we consider the update in \eqref{eq:gamma_mode} for a single parameter $\gamma_i$ while considering the estimates $\gamma^{[t]}_k$, $k\neq i$, and $\lambda^{[t]}$ as fixed quantities. In \ref{app:sbl_wi}, we show that the dependency of $\langle|w_i|^2\rangle^{[t]}$ on $\gamma^{[t]}_i$ is expressed as
\begin{align}
\langle|w_i|^2\rangle^{[t]} = \frac{(\gamma_i^{[t]})^2(s_i^{[t]}+|q_i^{[t]}|^2)+\gamma_i^{[t]}(s_i^{[t]})^2}{(\gamma_i^{[t]}+s_i^{[t]})^2}
\label{eq:wi}
\end{align}
with $s_i^{[t]} \triangleq \ev_i^\trans\Sigmamat_{-i}^{[t]}\ev_i$, $q_i^{[t]} \triangleq \lambda^{[t]}\ev_i^\trans\Sigmamat_{-i}^{[t]}\Hmat^\hermit\y$, $\Sigmamat_{-i}^{[t]}\triangleq (\lambda^{[t]} \Hmat^\hermit\Hmat+\sum_{k\neq i}(\gamma_k^{[t]})^{-1}\ev_k\ev_k^\trans)^{-1}$ and $\ev_i$ denoting an $N\times1$ vector of all zeros but $1$ at the $i$th position.  
By inserting \eqref{eq:wi} into \eqref{eq:gamma_mode}, we obtain an update expression of the form
\begin{align}
\gamma_i^{\text{new}} = \varphi_i^{[t]}(\gamma_i^{\text{old}})
\label{eq:fix_point_upd}
\end{align}
where the function $\varphi_i^{[t]}$ is parametrized by $\epsilon$, $\eta$, $s_i^{[t]}$, and $q_i^{[t]}$. Next, we want to explore the hypothetical behavior of the estimates of $\gamma_i$ that we would obtain by recursively applying $\varphi_i^{[t]}$ \emph{ad infinitum}. We do so by analyzing the existence of fixed points of the function $\varphi_i^{[t]}$. A fixed point $\gammatilde_i$ of $\varphi_i^{[t]}$ must fulfill
\begin{align}
\gammatilde_i = \varphi_i^{[t]}(\gammatilde_i) = \frac{\epsilon-\rho-1 + \sqrt{(\epsilon-\rho-1)^2+4\rho\eta\frac{\gammatilde_i^2(s_i+|q_i|^2)+\gammatilde_is_i^2}{(\gammatilde_i+s_i)^2}}}{2\eta}
\label{eq:fix_point_upd2}
\end{align}
where, for notational simplicity, we have dropped the iteration index for $s_i$ and $q_i$. By inspection of \eqref{eq:fix_point_upd2}, it is clear that $\gammatilde_i = 0$ is always a fixed point of $\varphi_i^{[t]}$ when $\epsilon\leq 1$. We look for other positive fixed points by solving \eqref{eq:fix_point_upd2}. These fixed points are solutions of the fourth order equation
\begin{align}
0 &= \gamma_i\Big(\eta\gamma_i^3+\gamma_i^2[2\eta s_i-(\epsilon-\rho-1)] \notag\\
&\quad+\gamma_i[\eta s_i^2 -2(\epsilon-\rho-1)s_i-\rho(s_i+|q_i|^2)]-(\epsilon-1)s_i^2\Big).
\label{eq:fourthorder}
\end{align}
Hence, if any strictly positive fixed point $\gammatilde_i$ of $\varphi_i^{[t]}$ exists, it must be a solution of the cubic equation
\begin{align}
0 &= \eta\gamma_i^3+\gamma_i^2[2\eta s_i-(\epsilon-\rho-1)] \notag\\
&\quad+\gamma_i[\eta s_i^2 -2(\epsilon-\rho-1)s_i-\rho(s_i+|q_i|^2)]-(\epsilon-1)s_i^2.
\label{eq:cubic}
\end{align}
As we show in \ref{app:sbl_ell}, the positive solutions of \eqref{eq:cubic} correspond, in fact, to the stationary points of \eqref{eq:sbl_cost} when all variables except $\gamma_i$ are kept fixed at their current estimates, i.e., of
\begin{align}
\ell_i^{[t]}(\gamma_i) \propto^{e} \log(p(\y|\gamma_i,\gam_{-i}^{[t]},\lambda^{[t]})p(\gamma_i)). 
\label{eq:lgammai}
\end{align}

Based on the above analysis, we formulate a new update rule for $\gamma_i$ at iteration $t~+~1$. Given the values of all estimates at iteration $t$, we calculate the fixed points of the corresponding function $\varphi_i^{[t]}$ by solving \eqref{eq:fourthorder}. Then
\begin{itemize}
\item if no strictly-positive fixed points of $\varphi_i^{[t]}$ exist, we set $\gamma_i^{[t+1]} = 0$, which, remember, is also a fixed point of $\varphi_i^{[t]}$.
\item if strictly-positive fixed points of $\varphi_i^{[t]}$ exist, we select the fixed point $\gammatilde_i$ which yields the largest value $\ell_i^{[t]}(\gammatilde_i)$ among all strictly positive fixed points. We then set $\gamma_i^{[t+1]} = \gammatilde_i$.
\end{itemize}
Note that the above selection criterion for $\gamma_i^{[t+1]}$ is a heuristic choice. In fact, we have no guarantee that, by iteratively applying the recurrent function $\varphi_i^{[t]}$, convergence to the selected fixed point will occur. This is likely to depend on the initialization $\gamma_i^{[t]}$. Moreover, when $\epsilon<1$, selecting a strictly-positive fixed point instead of 0 does not guarantee that the objective function \eqref{eq:sbl_cost} is increased, as \eqref{eq:lgammai} diverges to infinity when $\gamma_i$ tends to 0.\footnote{See the discussion in Section~\ref{sec:t1t2estimators}.}  With this selection, however, we hope to obtain an improved convergence speed at the expense of sacrificing the monotonicity property of the EM algorithm. The numerical results obtained with this heuristic choice, shown in Section~\ref{sec:simulations}, confirm the effectiveness of the approach.

Next we investigate the solutions of \eqref{eq:fourthorder} for different selections of $\epsilon$ and $\eta$. We show that for some particular selections of these parameters, the modified update of $\gamma_i^{[t+1]}$ coincides with the updates in the algorithms presented in \cite{Tipping2003} and \cite{Babacan2010}.
For brevity, we omit the algorithmic iteration index $t$ throughout the rest of the section.

\subsubsection{Fixed points for $0\leq\epsilon<1$ and $\eta\geq0$}

We consider an arbitrary value of $\epsilon$ in the range $0\leq\epsilon<1$. First, as $-(\epsilon-1)s_i^2 \geq 0$ for $\epsilon<1$, \eqref{eq:cubic} has at least one negative solution. If no positive solution exists we set $\gammahat_i =0$. If \eqref{eq:cubic} has at least one positive solution it is easily shown that it has exactly two, denoted by $\gamma_i^{(1)}$ and $\gamma_i^{(2)}$. If $\gamma_i^{(1)}=\gamma_i^{(2)}$ then this point is a saddle point of $\ell_i$ and therefore we set $\gammahat_i=0$. If $\gamma_i^{(2)}>\gamma_i^{(1)}$ then $\gammahat_i = \gamma_i^{(2)}$ or if $\gamma_i^{(1)}>\gamma_i^{(2)}$ then $\gammahat_i = \gamma_i^{(1)}$ (the proof is straightforward and is omitted). Thus, we always select the right-most positive solution.

For the special case $\epsilon=\eta =0$, i.e., when the mixing density coincides with the Jeffreys prior, \eqref{eq:cubic} reduces to a quadratic equation. It is easy to show that in this case either two positive solutions exist or none exists.  
 
\subsubsection{Fixed points for $\epsilon=1$ and $\eta=0$}

In this case the mixing density coincides with a constant improper prior, which leads to the same GSM model as used in RVM \cite{Tipping2001,Tipping2003,WipfRao04}. With this setting \eqref{eq:cubic} simplifies to 
\begin{align}
	\gammahat_i = |q_i|^2-s_i. 
\label{eq:gamma_fastrvm}
\end{align}
From \eqref{eq:gamma_fastrvm}, a positive solution of \eqref{eq:cubic} exists if and only if $|q_i|^2>s_i$. If this condition is not satisfied we set $\gammahat_i =0$. It is interesting to note that \eqref{eq:gamma_fastrvm} is independent of $\rho$ and thus is the same regardless of whether the signal model \eqref{eq:model} is real or complex. 

Next, we show the equivalence between \eqref{eq:gamma_fastrvm} and the corresponding update in Fast RVM \cite{Tipping2003}. In \cite{Tipping2003}, the estimate of $\gamma_i$ is computed as the maximizer of the marginal log-likelihood function $\ell_i(\gamma_i,\epsilon=1,\eta=0)$ in \eqref{eq:lgammai}. Hence, the estimate of $\gamma_i$ in \cite{Tipping2003} equals that in \eqref{eq:gamma_fastrvm}, because \eqref{eq:gamma_fastrvm} maximizes $\ell_i(\gamma_i,\epsilon=1,\eta=0)$.   
As the updates of $\muv$, $\Sigmamat$, and $\lambdahat$ are identical to those in Fast RVM the two algorithms coincide when $\epsilon=1$ and $\eta = 0$.

\subsubsection{Fixed points for $\epsilon=1$ and $\eta>0$}

In this case the mixing pdf coincides with an exponential pdf, so the GSM model is the same as that used in Fast Laplace \cite{Babacan2010}. The solution
\begin{align} 
	\gammahat_i= \frac{-(2\eta s_i +\rho)+\sqrt{\rho^2+4\rho\eta|q_i|^2}}{2\eta}
\label{eq:gamma_fastlaplace}
\end{align}
is positive if and only if $|q_i|^2-s_i > \eta s_i^2/\rho$ otherwise we set $\gammahat_i=0$. The case $\epsilon=1$ and $\rho=1/2$ corresponds to the GSM model of the Laplace prior for real weights. Obviously, \eqref{eq:gamma_fastlaplace} can also be used for complex weights, with $\rho=1$. Yet in this case the marginal prior for $\w$ is  no longer Laplacian, as showed in Section~\ref{sec:prior_models}, but some other sparsity-inducing member of the Bessel K density family. The estimate of $\gamma_i$ in Fast Laplace \cite{Babacan2010} is the maximizer of $\ell_i(\gamma_i,\epsilon=1,\eta)$ and, hence, is identical to the estimate in \eqref{eq:gamma_fastlaplace}. 

%

\subsection{Fast Sequential Inference Scheme} 


The modified update of $\gamma_i^{[t+1]}$, $i=1,\ldots,N$, described in Section~\ref{subsec:fastscheme} can be directly used to speed up the EM algorithm presented in Section~\ref{sec:sbl_em2}. With this modification, every time an estimate of a given $\gamma_i$ is set to zero, we remove the corresponding column vector $\h_i$ from the dictionary matrix $\Hmat$.
This effectively reduces the model complexity ``on the fly''. However, the first iterations still suffer from a high computational complexity due to the update \eqref{eq:wcov}. To avoid this, we follow the approach outlined in \cite[Sec.~4]{Tipping2003}, which consists of starting with an ``empty'' dictionary $\Hmat$ and incrementally filling the dictionary by possibly adding one column vector at each iteration of the algorithm. 
Specifically, at a given iteration of the algorithm, each $\gammahat_i$, $i=1,\ldots,N$, is computed  from \eqref{eq:fourthorder} and the one, say $\gammahat_{i'}$, that gives rise to the greatest increase in $\exp(\ell(\cdot))$ between two consecutive algorithmic iterations, is selected. Depending on the value of this $\gammahat_{i'}$, the corresponding vector $\h_{i'}$ is then added, deleted, or kept. The quantities $\Sigmamat$, $\muv$, and $\lambdahat$ are updated using \eqref{eq:wcov}, \eqref{eq:wmean}, and \eqref{eq:lambda_mode} together with $s_i$ and $q_i$, $i=1,\ldots,N$. 
If the estimate of $\lambda$ is not updated between two consecutive iterations, $\Sigmamat$, $\muv$, $s_i$, and $q_i$ can be updated efficiently using the update procedures proposed in \cite{Tipping2003,ShutinFastRVM}. 

We refer to the above sequential algorithm as \textit{Fast-BesselK}.

%

%
\section{Numerical Results \label{sec:simulations}}
In this section we analyze the performance of the Fast-BesselK algorithm proposed in Section~\ref{sec:sbl}. 
The purpose is to characterize the impact of the prior model on the performance of the iterative algorithm in terms of MSE, sparseness of $\what$, and convergence speed. Section~\ref{sec:sbl} shows that Fast-RVM \cite{Tipping2003}, Fast-Laplace \cite{Babacan2010}, and Fast-BesselK are all instances of the same greedy inference scheme each algorithm resulting from a particular selection of the parameters of the mixing (gamma) pdf. Hence, by comparing the performances of these algorithms we can draw conclusions on the sparsity-inducing property of their respective prior models.\footnote{Naturally, the practical implementation of the inference schemes also impacts the performance. For the subsequent analysis, Fast-RVM, Fast-Laplace, and Fast-BesselK are all implemented based on the Matlab-code for Fast-RVM located at \url{http://people.ee.duke.edu/~lcarin/BCS.html}.} 

\subsection{Simulation Scenarios and Performance Metrics}

The performance of the considered sparse algorithms (see Section~\ref{subsec:algorithms}) is evaluated by means of Monte Carlo simulations. In order to test the algorithms on a realistic benchmark, we use a random $M\times N$ dictionary matrix $\Hmat$, with $M=100$ and $N=256$, whose entries are iid zero-mean complex symmetric Gaussian random variables with variance $M^{-1}$. The weight vector $\w$ has $K$ nonzero entries with associated indices uniformly drawn without repetition from the set $\{1,2,\ldots,N\}$. The set of these indices together with its cardinality $K$ are unknown to the algorithms. The nonzero entries in $\w$ are independent and drawn from a zero-mean circular-symmetric complex Gaussian distribution with unit variance. Other distributions for the entries in $\w$ are considered at the end of this section. All reported performance curves are computed based on a total of $1000$ Monte Carlo trials. For each trial, new realizations of the dictionary matrix $\Hmat$, the vector $\w$, and the random perturbation vector $\n$ are drawn independently.\footnote{In this paper we have not included an investigation on a specific application. We refer to the work \cite{Pedersen2013} where such a performance assessment is made.} 

All numerical investigations where replicated for an equivalent real-valued signal model. Due to space limitations, we do not include the results of these studies in this contribution, as most of the conclusions are similar to those drawn from the complex-valued signal model. We will, however, shortly discuss the relation between the performance of the estimators for real and complex models at the end of this section.

The performance is evaluated with respect to the following metrics: 
\begin{align}
	&\textit{normalized mean-squared error}:
  \; \mathrm{NMSE} \triangleq \langle\|\what-\w\|_2^2\rangle/\langle\|\w\|_2^2\rangle. \notag\\
	&\textit{support error rate}
	\; \triangleq \#\{\{i:\hat{w}_i = 0\;\mathrm{and}\; w_i \neq 0\} \cup \{i: \hat{w}_i \neq 0\;\mathrm{and}\; w_i = 0 \} \}/N \notag.
\end{align}
We also report the convergence speed, measured in terms of the number of algorithmic iterations used, of the Bayesian inference methods as they share the same computational complexity. 

\subsection{Inference Algorithms Considered \label{subsec:algorithms}}

The proposed Fast-BesselK algorithm is tested with two settings for $\epsilon$ and $\eta$: 
\begin{itemize}
\item Fast-BesselK($\epsilon=0$): we set $\epsilon =0$ and $\eta=0$ corresponding to the use of the Jeffreys prior as mixing density.\footnote{We also considered Fast-BesselK with $\epsilon =0$ and $\eta=1$. However, this setting led to similar performance to Fast-BesselK($\epsilon=0,\eta=0$).} 
\item Fast-BesselK($\epsilon=0.5$): we set $\epsilon=0.5$ and $\eta=1$.
\end{itemize}
Instead of selecting a particular value of $\eta$, we could have included this parameter in the inference framework as done in \cite{Babacan2010}. Our investigations, however, show that for $\epsilon<<1$ the performance of Fast-BesselK becomes largely independent of the choice of $\eta$, and we have therefore simply selected $\eta=1$.\footnote{If the Fast-BesselK is implemented with a ``top-down'' approach (starting out with the full dictionary $\Hmat$) including individual rate parameters $\eta_i$ for each $w_i$, $i=1,\ldots,N$, may be beneficial \cite{Pedersen2012}.} 


The performance of Fast-BesselK is contrasted with the state-of-the-art sparse estimators listed below:
\begin{enumerate}
	\item Fast-RVM \cite{Tipping2003, Ji2008}: 
	is equivalent to Fast-BesselK with $\epsilon=1$ and $\eta=0$ (see Section~\ref{sec:sbl}).\footnote{The software is available on-line at \url{http://people.ee.duke.edu/~lcarin/BCS.html}.}
	\item Fast-Laplace \cite{Babacan2010}: 
	is equivalent to Fast-BesselK with $\epsilon=1$ when including the update for $\eta$ in \cite{Babacan2010} (see Section~\ref{sec:sbl}).\footnote{The software is available on-line at \url{http://ivpl.eecs.northwestern.edu/}.}
	\item OMP, see e.g., \cite{Tropp2004}: OMP terminates when the greedy algorithm has included $K+10$ column vectors in $\Hmat$. We empirically observed that this choice induces a better NMSE performance than when including $K$ columns only.
	\item SpaRSA \cite{Wright2009}: the sparse reconstruction by separable approximation (SpaRSA) algorithm for solving the LASSO cost function. Following \cite{Wright2009}, we use the adaptive continuation procedure for the regularization $\kappa$ of the $\ell_1$-norm penalty in the LASSO cost function. Here SpaRSA repeatedly solves the LASSO cost function with decreasing values for $\kappa$ until a minimum value of $\kappa$ is reached. The minimum value of $\kappa$ is set through training. Specifically, we run 50 Monte Carlo trials for each specific settings of $M$, $N$, $K$, and SNR value. We then choose the value of $\kappa$ from a set of 50 candidate values in the range $[0.001\|\Hmat^\hermit y\|_\infty,0.1\|\Hmat^\hermit y \|_\infty]$ that leads to the smallest error $\|\w-\what\|_2^2$.\footnote{The software is available on-line at \url{http://www.lx.it.pt/~mtf/SpaRSA/}.}
\end{enumerate} 
For brevity, we refer to Fast-RVM, Fast-Laplace, and Fast-BesselK as Bayesian algorithms. We initialize these algorithms as outlined in \cite[Sec.~4]{Tipping2003}. They stop when either the stopping criterion $\|\muvhat^{[t+1]}-\muvhat^{[t]}\|_\infty \leq 10^{-8}$ is fulfilled or the number of iterations has reached its max limit set to 1000.

As a reference, we also consider the performance of the oracle estimator of $\w$ \cite{Candes2007} that ``knows'' the support of $\w$, denoted by $\supp(\w) \triangleq \{ i : w_i \neq 0 \}$. The oracle estimate reads  
\begin{align}
	\hat{\w}_{\mathrm{o}}(\y) = 
\left\{ \begin{array}{*{4}l} &\hspace{-10pt} (\Hmat^\hermit_{\mathrm{o}}\Hmat_{\mathrm{o}})^{-1}\Hmat^\hermit_{\mathrm{o}}\y &, \;\; \text{on} \;\; \mathrm{supp}(\w)  \\
&\hspace{-10pt} 0  &, \;\; \text{elsewhere},
\label{eq:oracle}
\end{array} \right. 
\end{align} 
where $\Hmat_{\mathrm{o}}$ is the $M\times K$ dictionary matrix constructed from those columns of $\Hmat$ that correspond to the nonzero entries in $\w$.

\subsection{Performance Comparison \label{subsec:performance}}

As our analysis in Section~\ref{sec:prior_models} shows, the noise precision $\lambda$ greatly impacts the sparsity property of the \tii{} penalty. We therefore investigate the impact of this parameter on the algorithms. First, we assume this quantity to be known to the Bayesian algorithms. Note that SpaRSA and OMP do not estimate $\lambda$. In a next step, this parameter is considered unknown and estimated by the Bayesian algorithms.

\subsubsection{Performance versus SNR} 

\begin{figure*}[!t]
\centering
\centerline{
\subfigure[]{\includegraphics[width=0.35\linewidth]{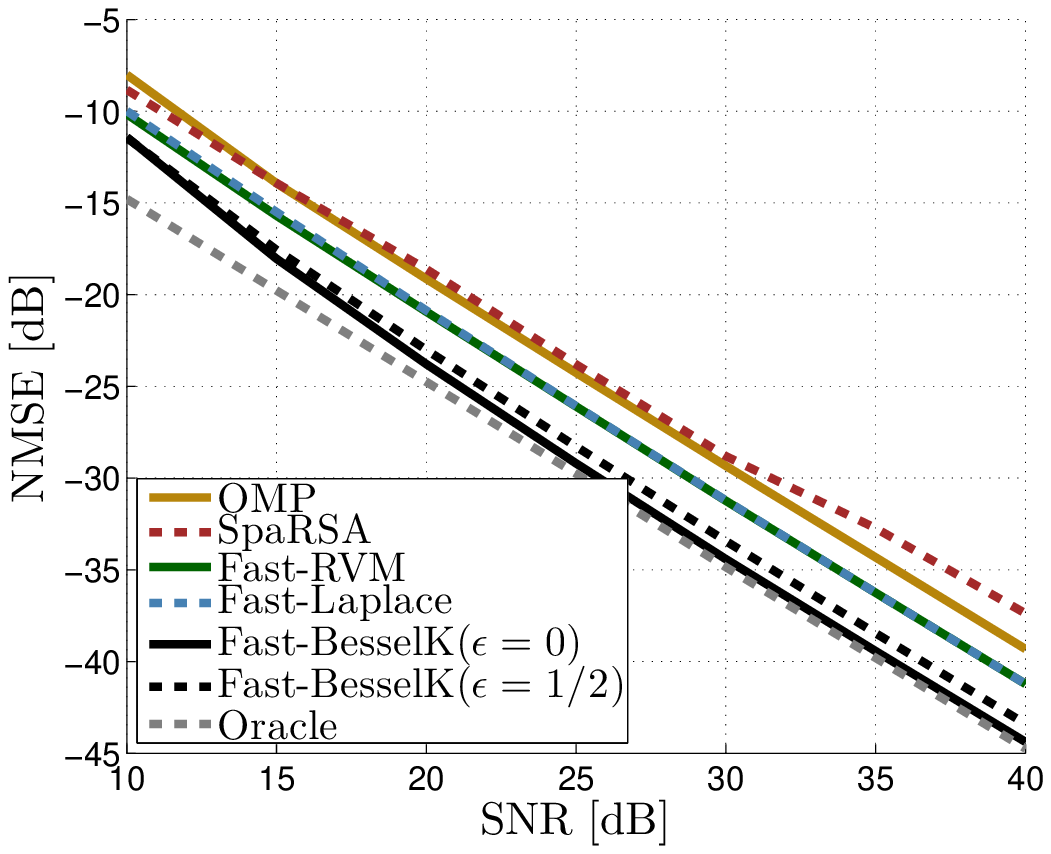}}
\subfigure[]{\includegraphics[width=0.35\linewidth]{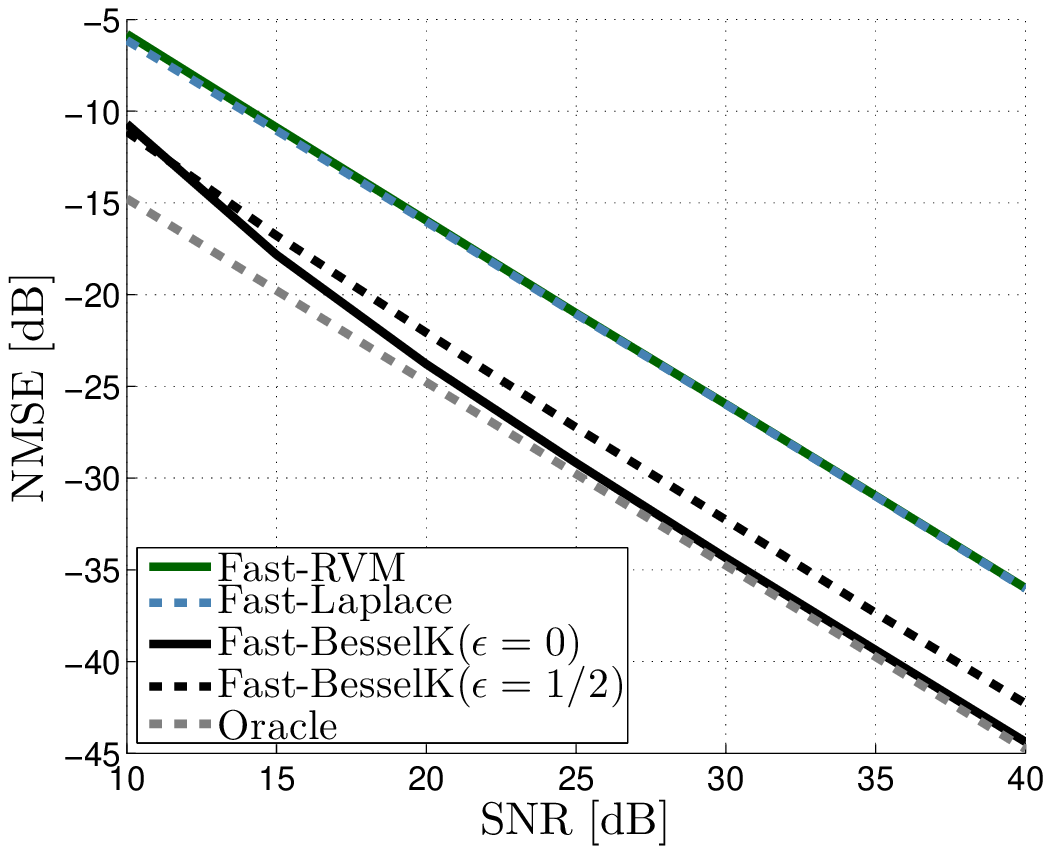}}
}
\centerline{
\subfigure[]{\includegraphics[width=0.35\linewidth]{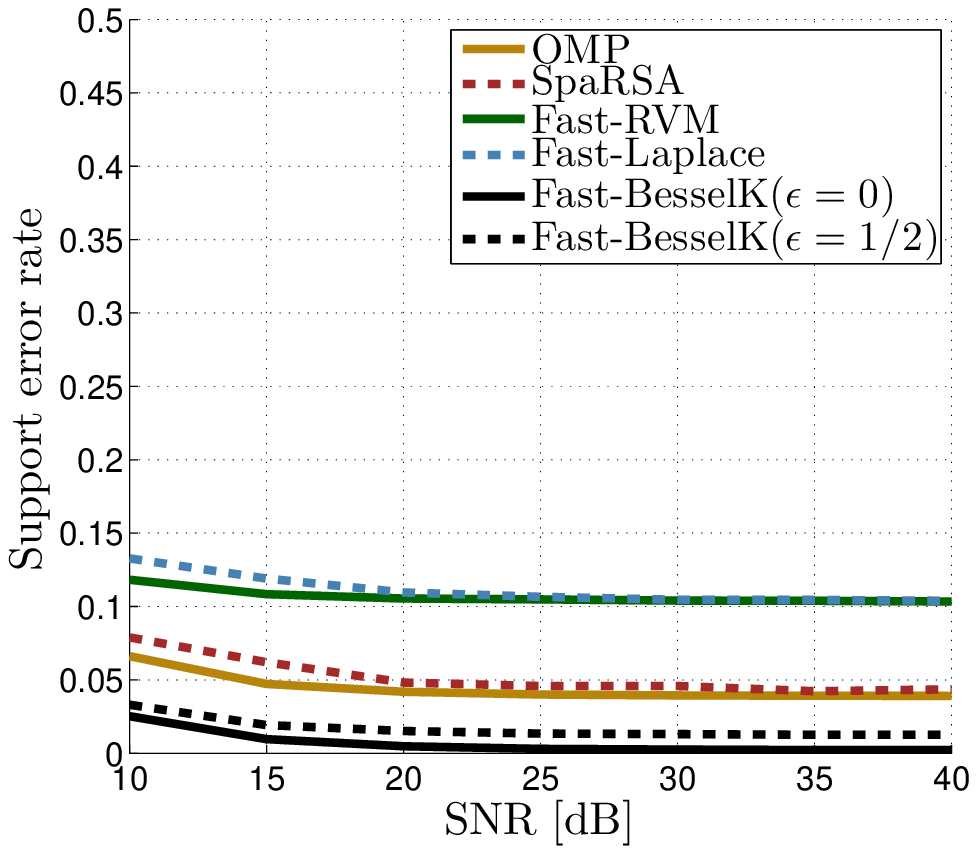}}
\subfigure[]{\includegraphics[width=0.35\linewidth]{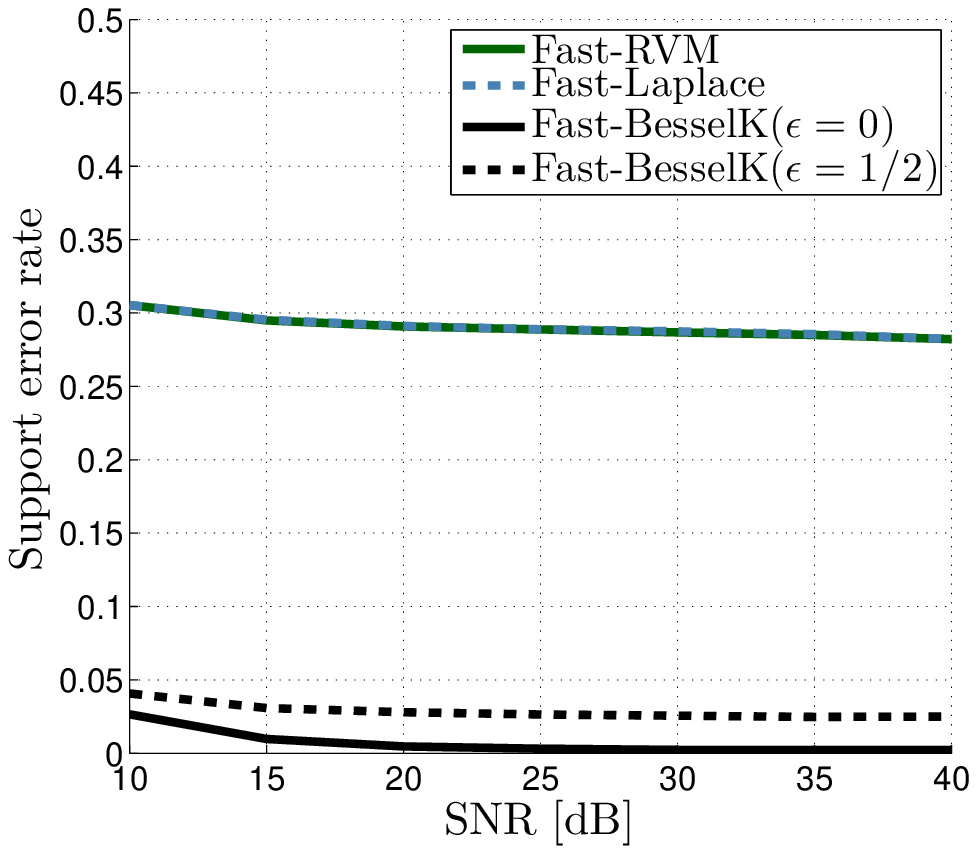}}
}
\caption{Performance versus SNR when $\lambda$ is known ((a), (c)) and $\lambda$ is unknown and estimated ((b), (d)). Selected system parameter settings: $M = 100$, $N = 256$, and $K = 25$. \label{fig:SNR}}
\end{figure*}

The goal of this investigation is to evaluate whether the algorithms can achieve sparse and accurate estimates in conditions of low and medium SNR. In these simulations, we set $K=25$. In \Fig~\ref{fig:SNR}(a) and \Fig~\ref{fig:SNR}(c), $\lambda$ is known by the Bayesian algorithms. \Fig~\ref{fig:SNR}(a) shows that Fast-BesselK($\epsilon=0$) and Fast-BesselK($\epsilon=0.5$) achieve the lowest NMSE among the algorithms across the whole SNR range. Their performance is close to that of the oracle estimator in the high SNR regime, i.e., above 20 dB. These algorithms also achieve the lowest support error rate across the whole SNR range with a value close to zero as shown in \Fig~\ref{fig:SNR}(c).

\begin{figure*}[!t]
\centering
\centerline{
\subfigure[]{\includegraphics[width=0.35\linewidth]{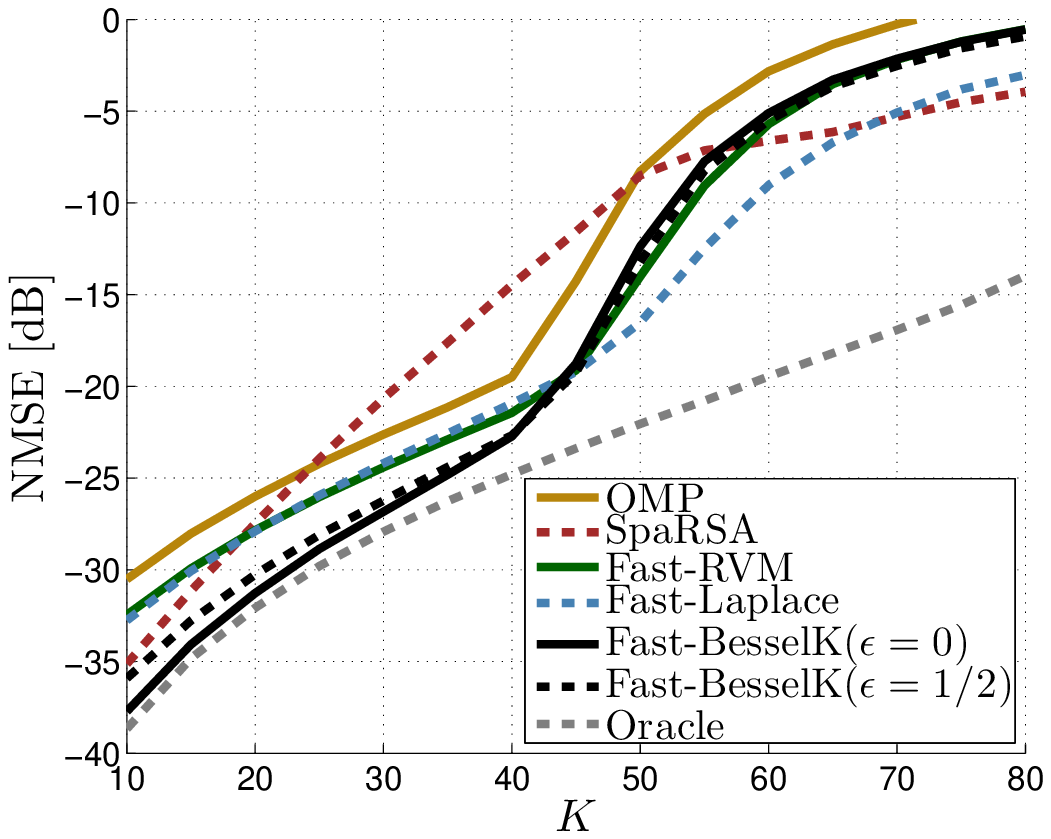}}
\subfigure[]{\includegraphics[width=0.35\linewidth]{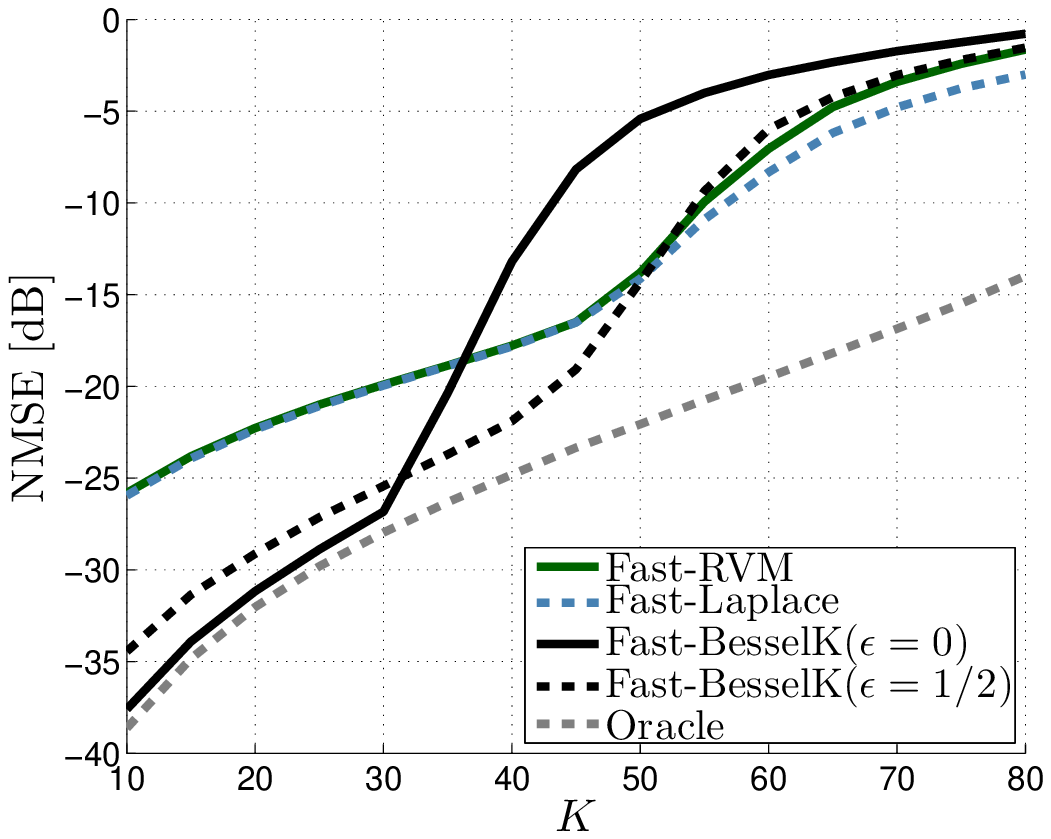}}
}
\centerline{
\subfigure[]{\includegraphics[width=0.35\linewidth]{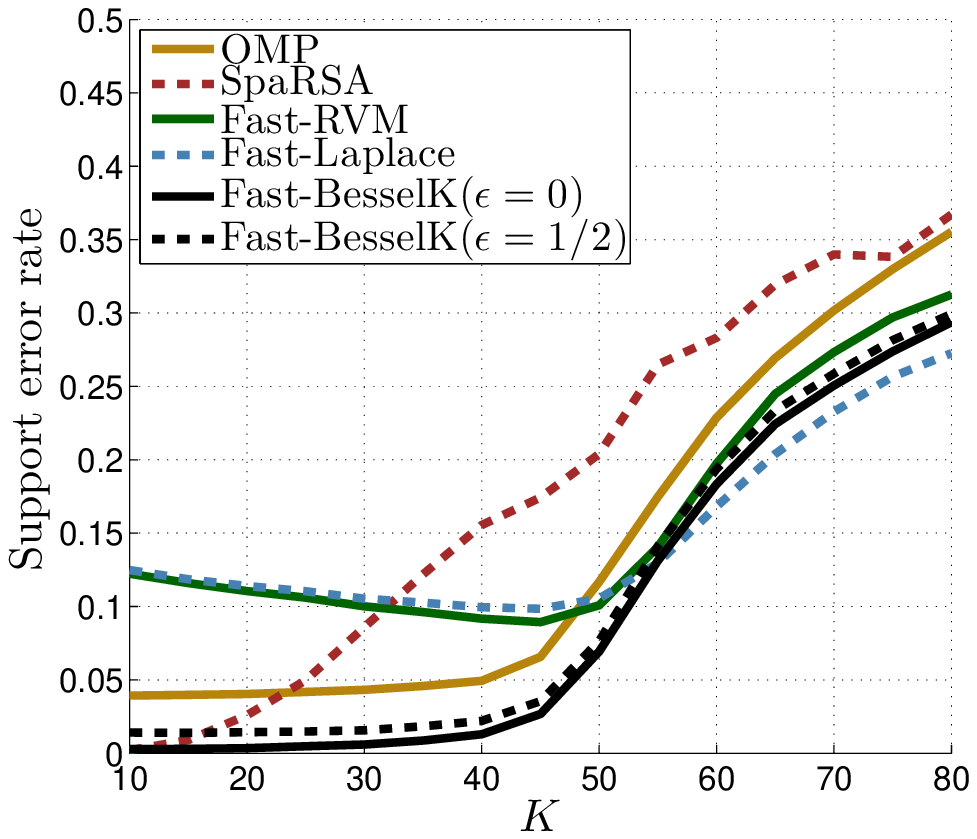}}
\subfigure[]{\includegraphics[width=0.35\linewidth]{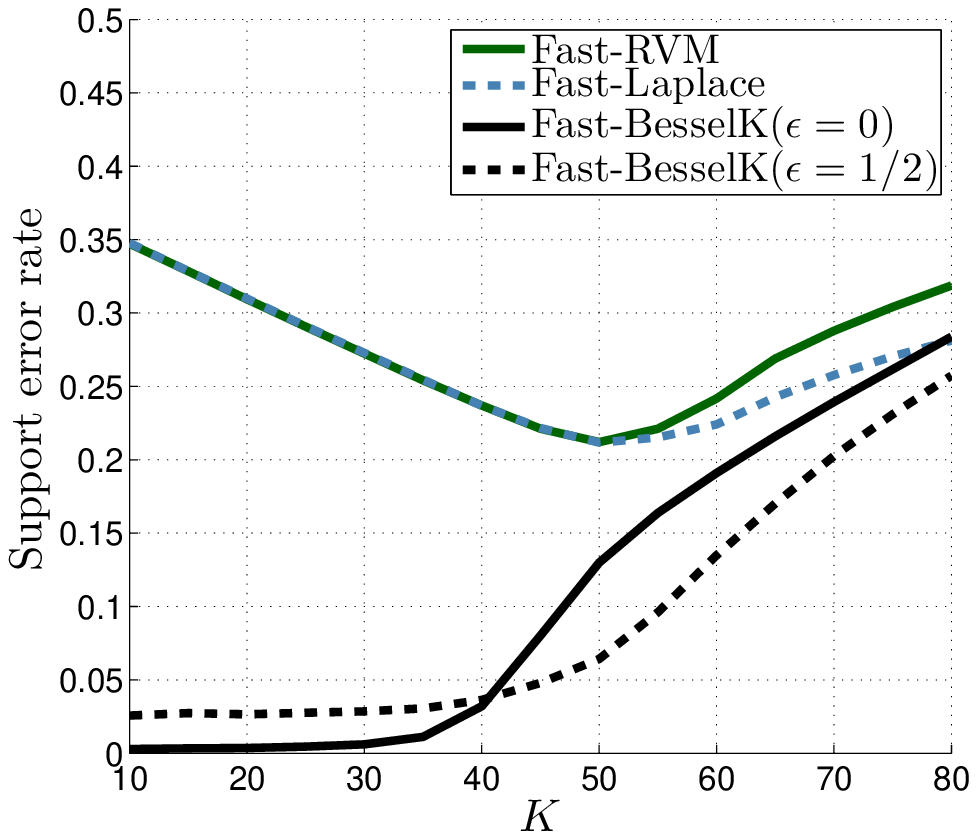}}
}
\caption{Performance versus $K$ at 20 dB SNR when $\lambda$ is known ((a), (c)) and $\lambda$ is unknown and estimated ((b), (d)). Selected system parameter settings: $M = 100$, $N = 256$, and $K = 25$.\label{fig:K}}
\end{figure*}

We repeat the investigation for the Bayesian algorithms but this time with the noise precision $\lambda$ unknown and being estimated alongside $\w$ and $\gam$ using \eqref{eq:lambda_mode}. The estimate $\lambdahat$ is updated at every third iteration. We observe a significant performance degradation in NMSE and support error rate for Fast-RVM and Fast-Laplace in \Fig~\ref{fig:SNR}(b) and \Fig~\ref{fig:SNR}(d). The reason is that Fast-RVM and Fast-Laplace heavily overestimate $\lambda$, thus, $K$ is overestimated as well (results not shown).\footnote{In some cases, the sequence of estimates of $\lambda$ produced by Fast-RVM and Fast-Laplace did not convergence. Due to this, a maximum of value of $10^8$ was set for $\lambdahat$.} Consequently, the support error rate and NMSE is high. In contrast, the Fast-BesselK algorithms perform essentially the same as when $\lambda$ is known.

In summary, the results presented in \Fig~\ref{fig:SNR} corroborate the significant impact of the estimation of the noise precision on the performance of the Fast Bayesian algorithms. When $\lambda$ is known, all algorithms achieve an acceptable performance, both in terms of NMSE and support error rate. However, when $\lambda$ is unknown and estimated by the algorithms, only Fast-BesselK is able to produce accurate estimates of this parameter, resulting in greatly improved performance as compared to Fast-Laplace and Fast-RVM. This is an important result as, in many applications, the noise precision parameter is not known in advance and, hence, needs to be estimated.  

\subsubsection{Performance versus $K$} 
 
We fix the SNR at 20 dB and compare the performance of the algorithms versus the number $K$ of nonzero entries in $\w$. In \Fig~\ref{fig:K}(a) and \Fig~\ref{fig:K}(c)  we assume $\lambda$ to be known to the Bayesian algorithms. The NMSE curves in \Fig~\ref{fig:K}(a) show that when $K\leq40$ the algorithms achieve an accurate reconstruction of $\w$. Fast-BesselK($\epsilon=0$) and Fast-BesselK($\epsilon=0.5$) yield the lowest NMSE which turns out to be close to that of the oracle estimator. In this range, these algorithms exhibit a support error rate close to zero as depicted in \Fig~\ref{fig:K}(c).

When $\lambda$ is estimated the NMSE and support error rate performance achieved by Fast-RVM and Fast-Laplace degrade as depicted in \Fig~\ref{fig:K}(c) and \Fig~\ref{fig:K}(d). Fast-BesselK($\epsilon=0$) achieves the lowest NMSE but only for $K\leq30$, as it only accurately estimates $\lambda$ in this range. Consequently, its support error rate decreases for $K>30$. In turn, Fast-BesselK($\epsilon=0.5$) achieves similar performance to when $\lambda$ is known. Hence, the selection of $\epsilon=0.5$ seems to be a good trade-off between achieved sparseness and reconstruction error.

\begin{figure*}[!t]
\centering
\centerline{
\subfigure[]{\includegraphics[width=0.35\linewidth]{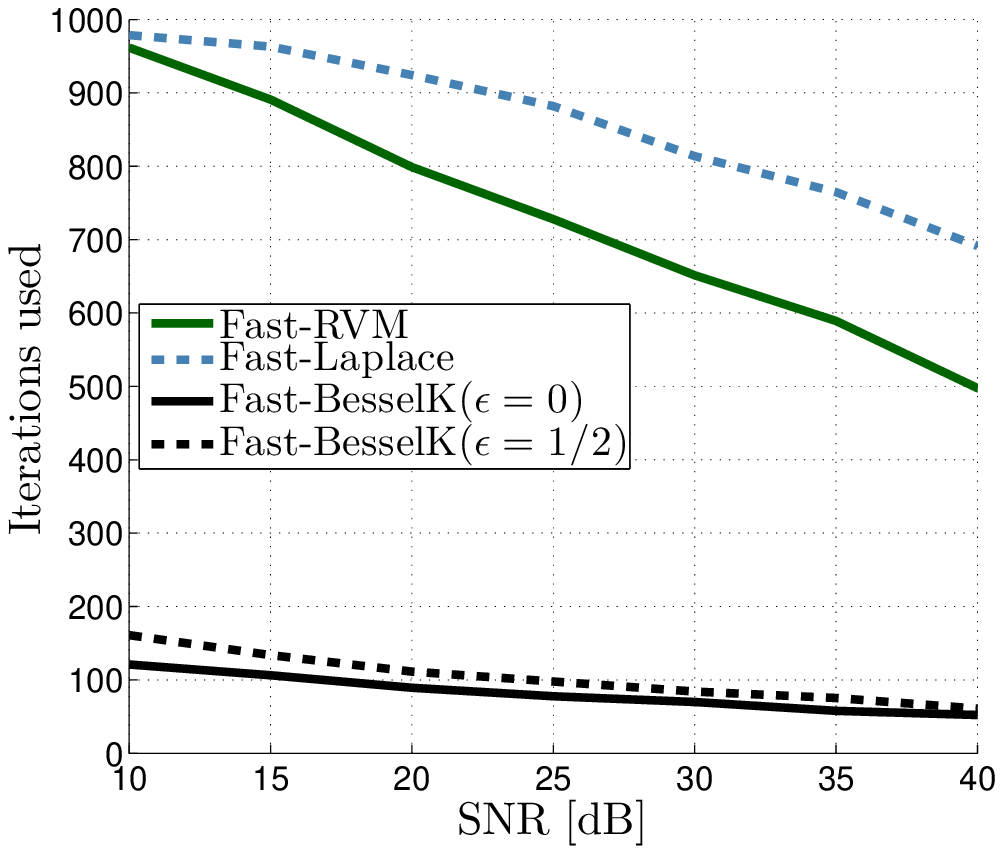}}
\subfigure[]{\includegraphics[width=0.35\linewidth]{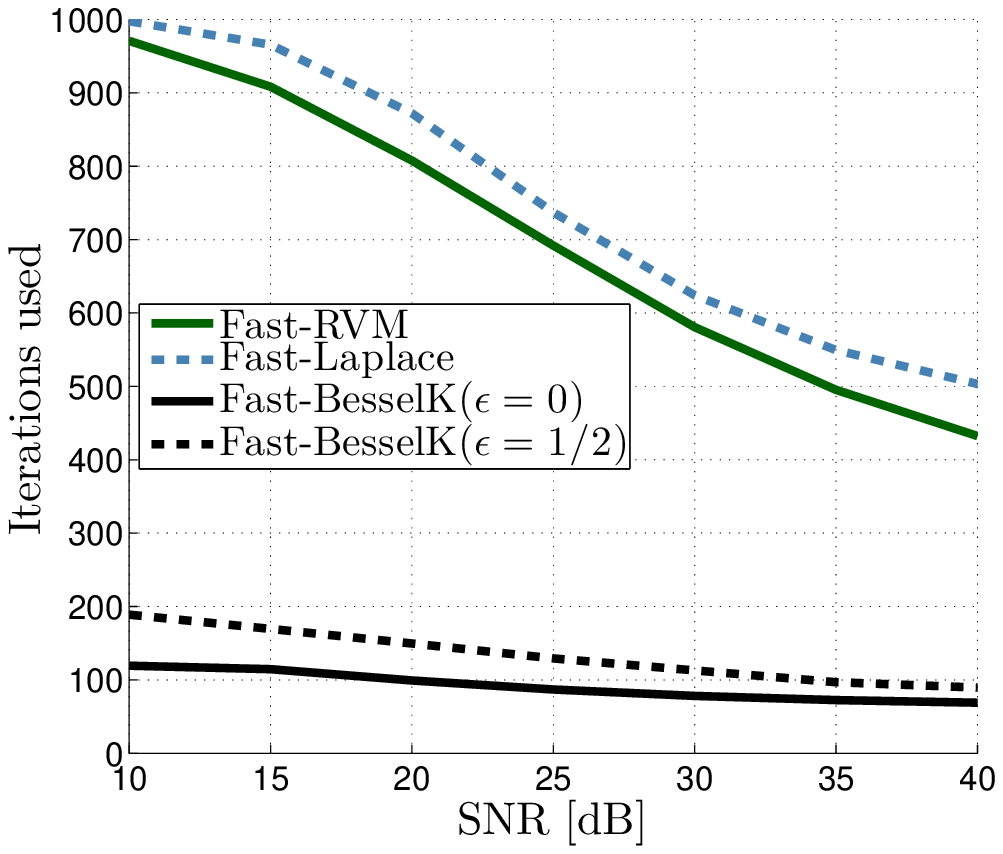}}
}
\centerline{
\subfigure[]{\includegraphics[width=0.35\linewidth]{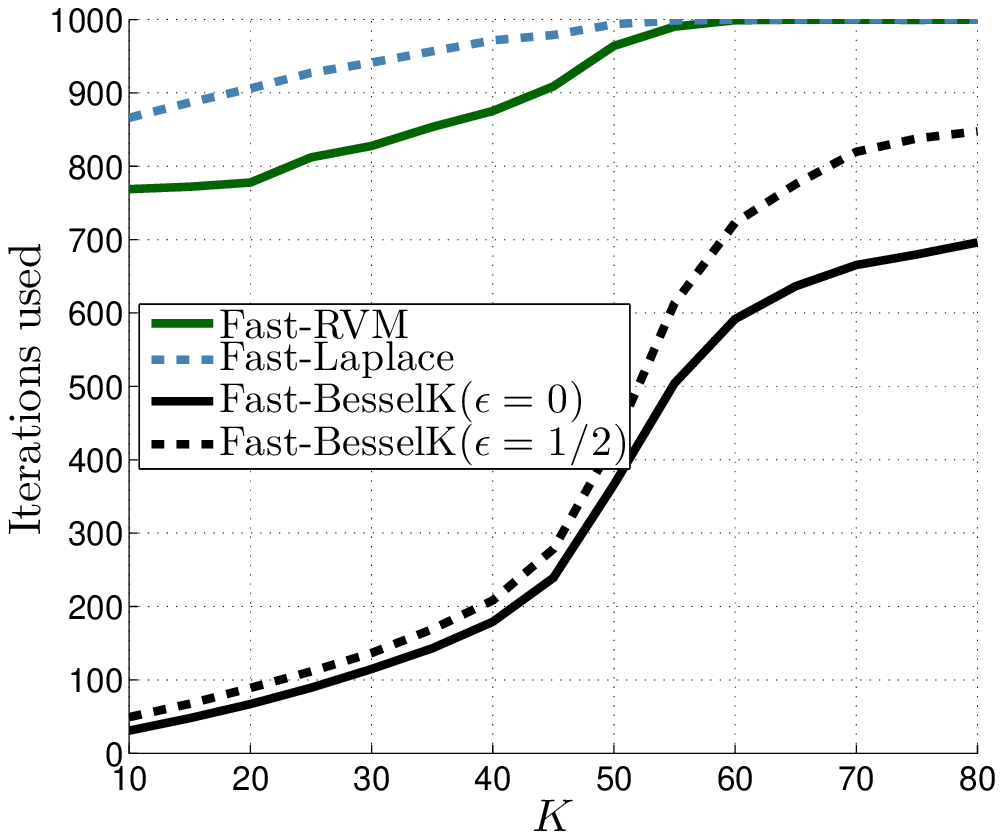}}
\subfigure[]{\includegraphics[width=0.35\linewidth]{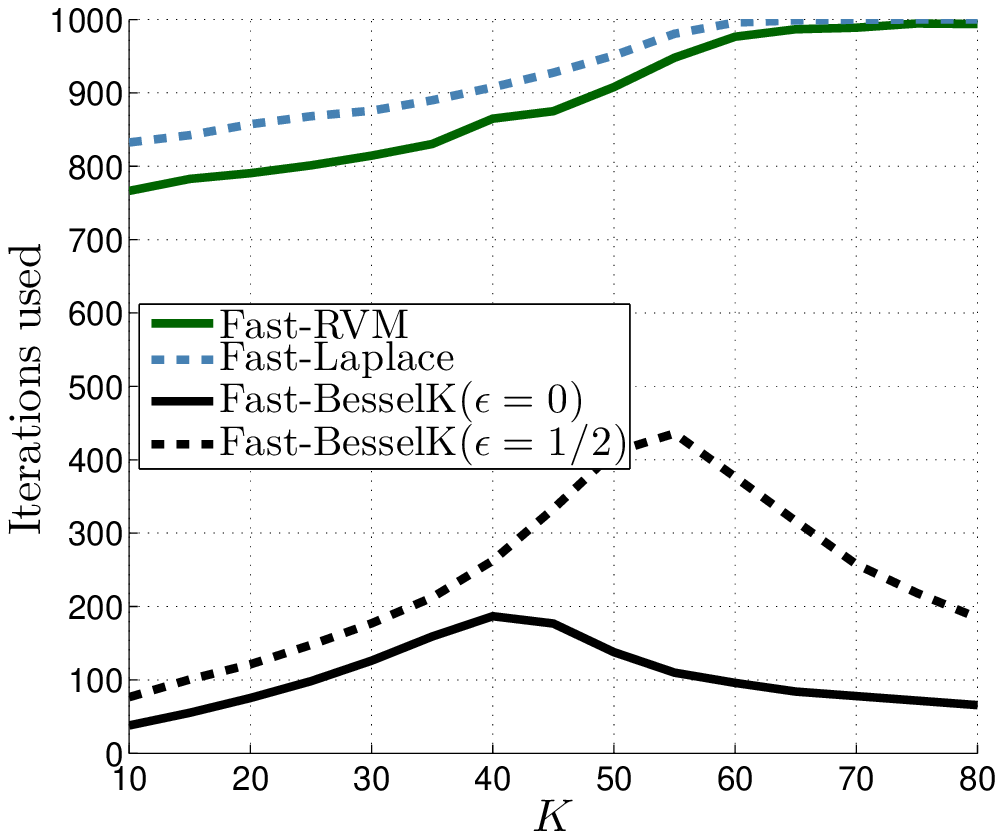}}
}
\caption{Number of used iterations versus SNR and $K$ when $\lambda$ is known ((a), (c)) and $\lambda$ is unknown and estimated ((b), (d)). Selected system parameter settings: $M = 100$, $N = 256$. In ((a), (b)) $K = 25$ and in ((c), (d)) the SNR is fixed at 20 dB. \label{fig:speed_SNRK}}
\end{figure*}

\subsubsection{Number of performed algorithmic iterations}

We evaluate the convergence speed for the Bayesian algorithms in terms of the number of performed algorithmic iterations. \Fig~\ref{fig:speed_SNRK} reports the number of algorithmic iterations until either the stopping criterion is fulfilled or the number of iterations has reached its max limit of 1000 (see Section~\ref{subsec:algorithms}) versus SNR and $K$. The Fast-BesselK algorithms perform significantly less number of iterations across the whole SNR range as compared to Fast-RVM and Fast-Laplace, especially in low to medium SNR as seen from \Fig~\ref{fig:speed_SNRK}(a) and \Fig~\ref{fig:speed_SNRK}(b). The same superior performance is observed when $K$ is varied in \Fig~\ref{fig:speed_SNRK}(c) and \Fig~\ref{fig:speed_SNRK}(d). Notice that the iteration count of greedy algorithms inherently depends on $K$. As Fast-RVM and Fast-Laplace tend to heavily overestimate $K$, they inevitably require a larger number of iterations than algorithms achieving sparser estimates. The Fast-BesselK algorithms exhibit a modest increase of used iterations when $K\leq40$ as they achieve good reconstruction error in this range, see \Fig~\ref{fig:K}. When $K\geq40$, the different performance behavior for Fast-BesselK in \Fig~\ref{fig:speed_SNRK}(c) and \Fig~\ref{fig:speed_SNRK}(d) is attributed to the fact that Fast-BesselK significantly underestimates $\lambda$ in this range. In this case, the penalty $q_{II}(\w)$ has a high impact on the estimate $\what$, which leads to a very sparse estimate $\what$ and, thus, a low number of algorithmic iterations.

\subsubsection{Performance versus different distributions of the nonzero entries in $\w$}

\begin{figure*}[!t]
\centering
\centerline{
\subfigure[]{\includegraphics[width=0.35\linewidth]{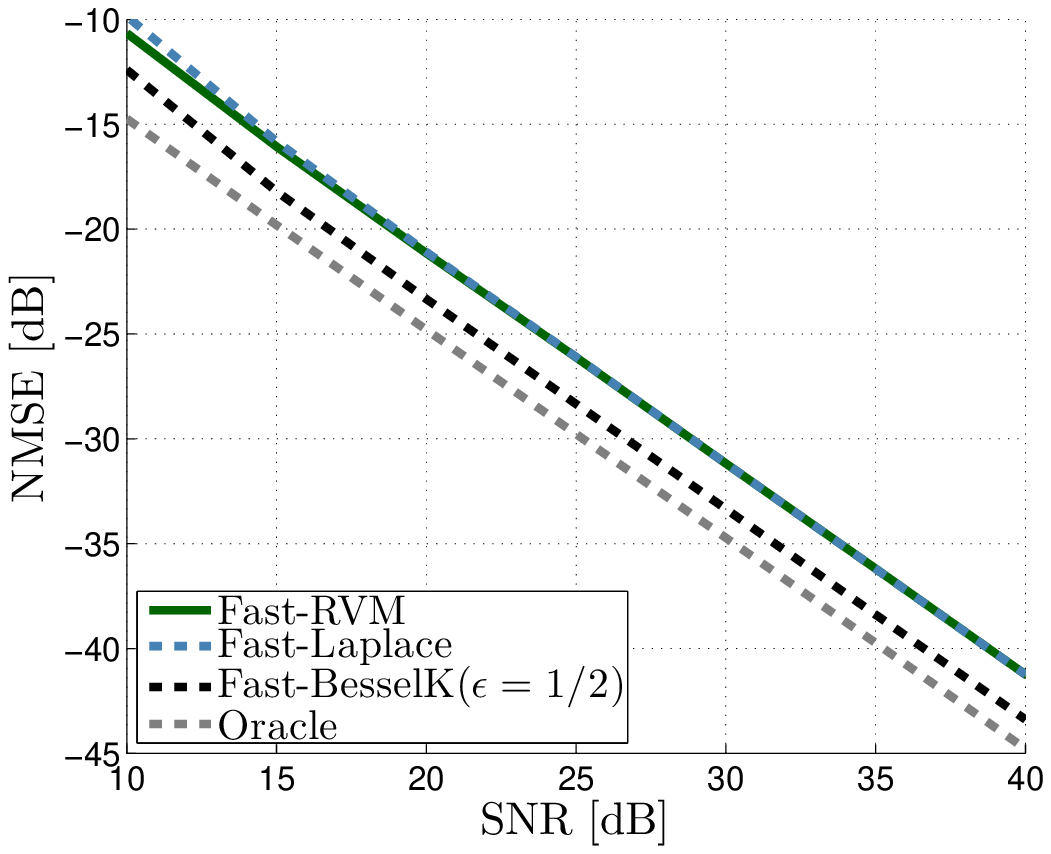}}
\subfigure[]{\includegraphics[width=0.35\linewidth]{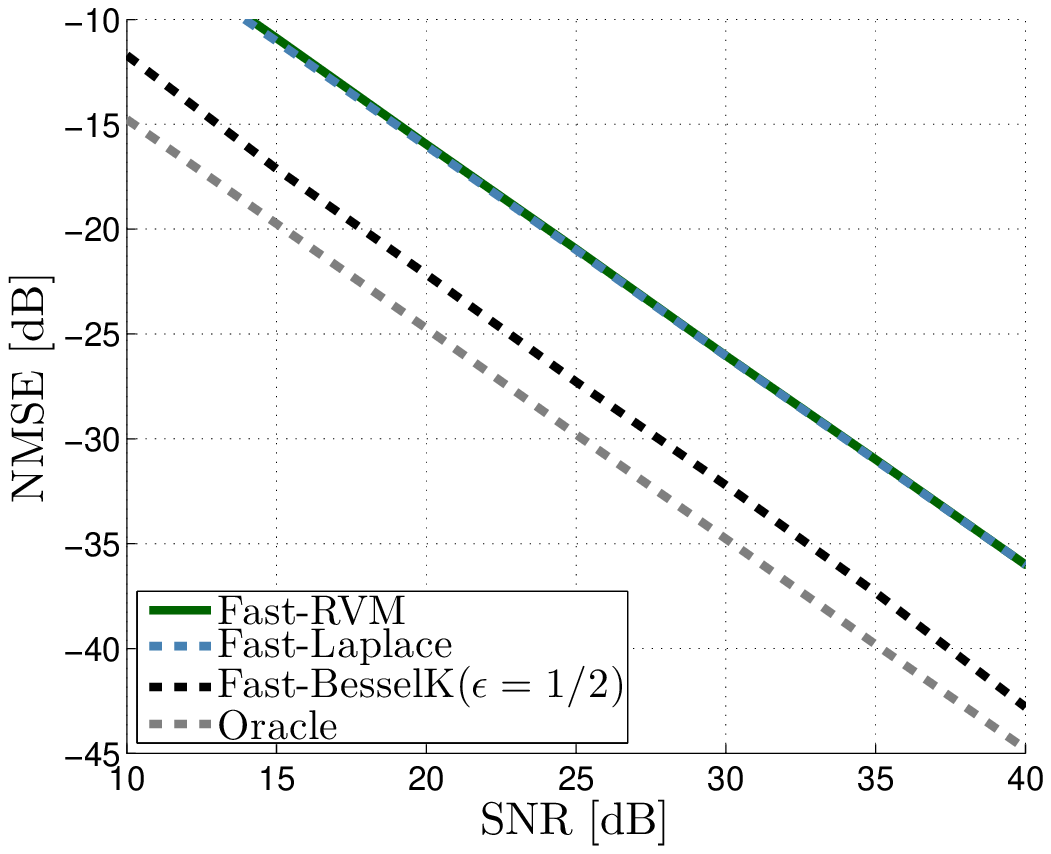}}
}
\centerline{
\subfigure[]{\includegraphics[width=0.35\linewidth]{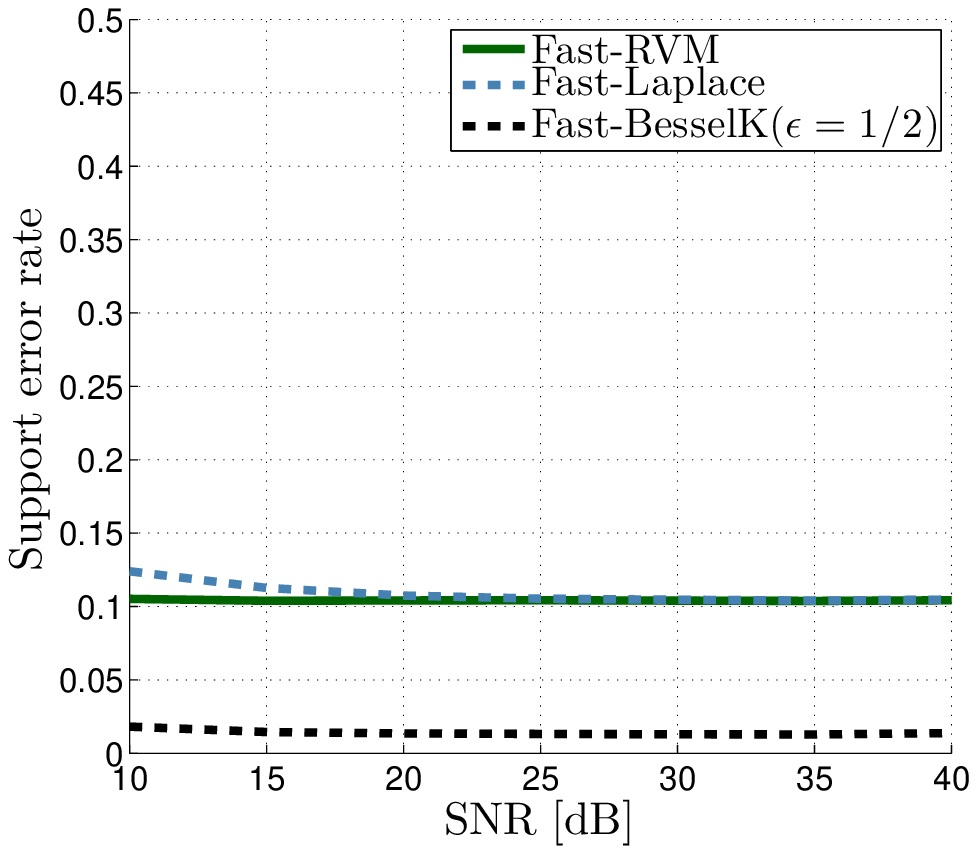}}
\subfigure[]{\includegraphics[width=0.35\linewidth]{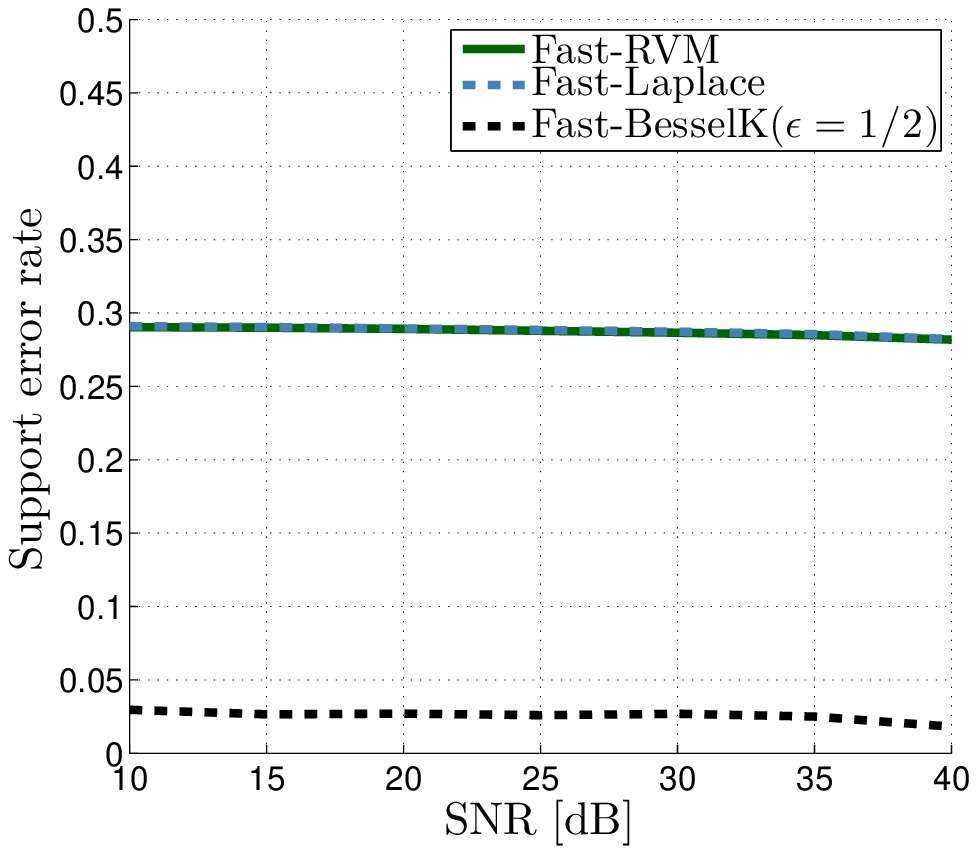}}
}
\caption{Performance versus SNR when $\lambda$ is known ((a), (c)) and $\lambda$ is unknown and estimated ((b), (d)). The nonzero entries in $\w$ are complex uniformly distributed. Selected system parameter settings: $M = 100$, $N = 256$, and $K = 25$.\label{fig:uniform}}
\end{figure*}

\begin{figure*}[!t]
\centering
\centerline{
\subfigure[]{\includegraphics[width=0.35\linewidth]{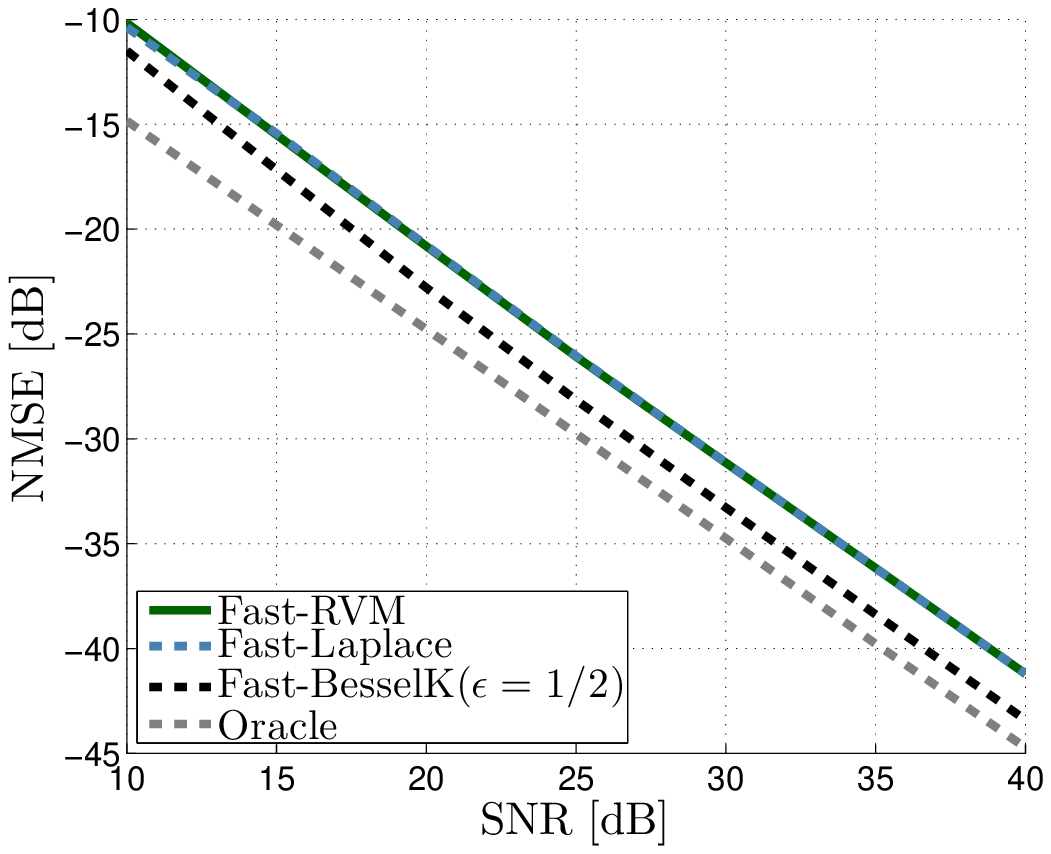}}
\subfigure[]{\includegraphics[width=0.35\linewidth]{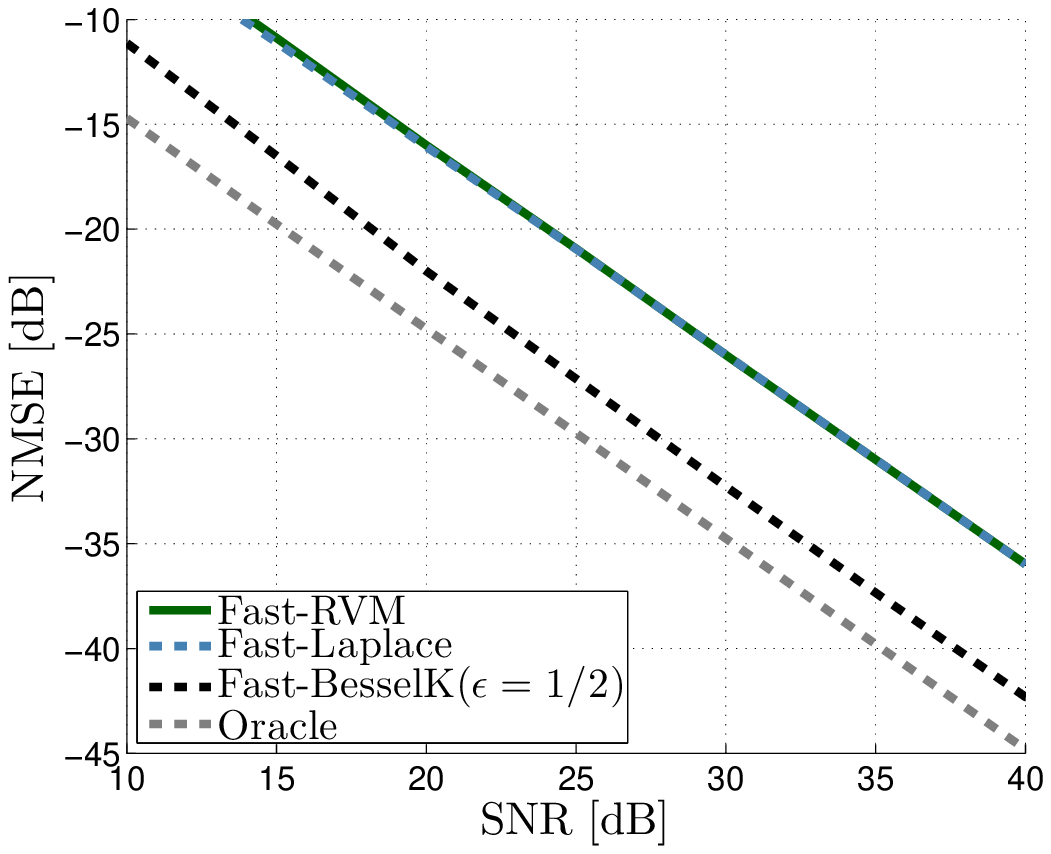}}
}
\centerline{
\subfigure[]{\includegraphics[width=0.35\linewidth]{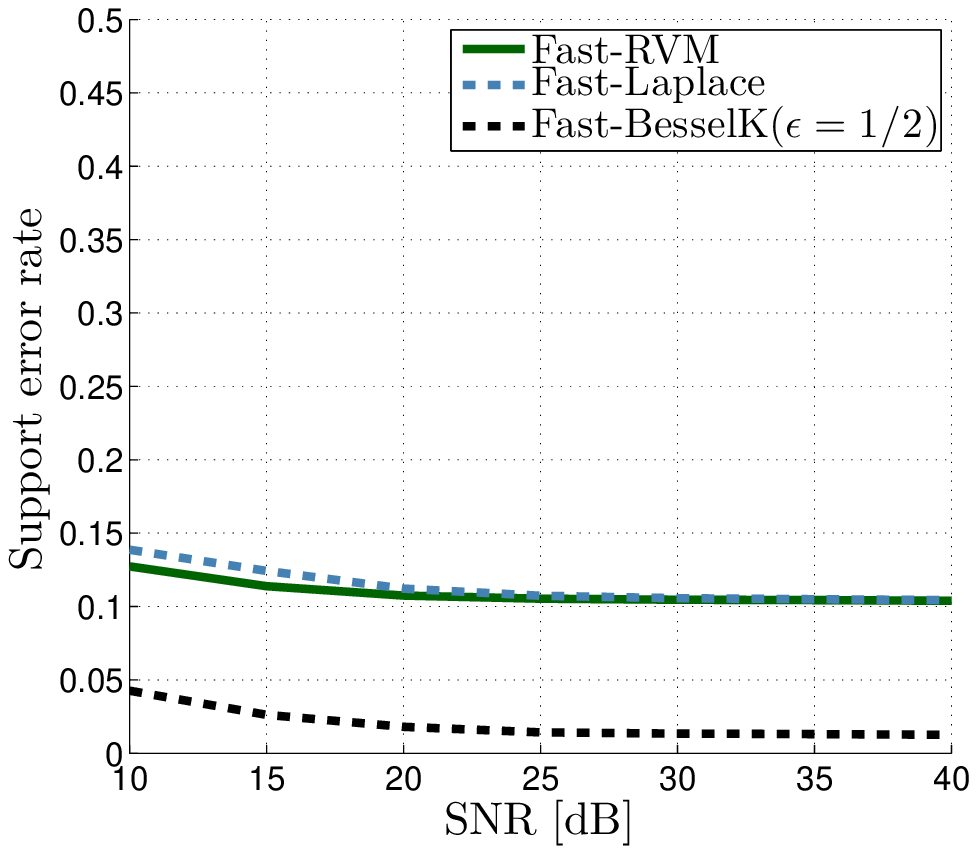}}
\subfigure[]{\includegraphics[width=0.35\linewidth]{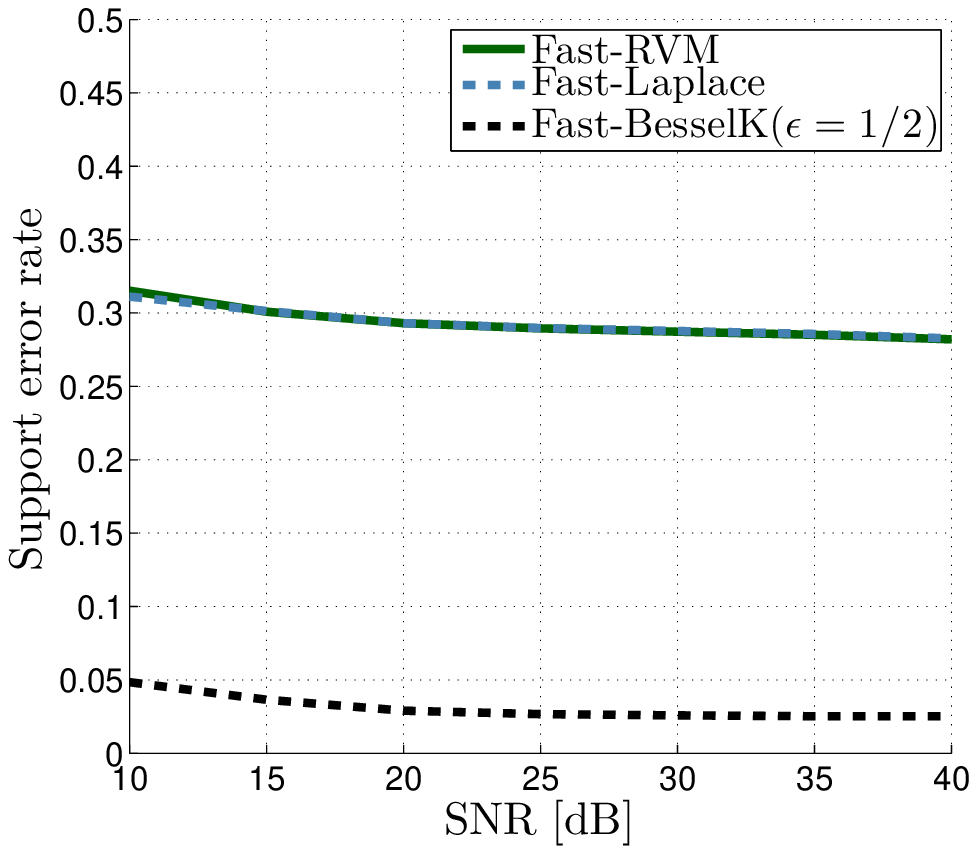}}
}
\caption{Performance versus SNR when $\lambda$ is known ((a), (c)) and $\lambda$ is unknown and estimated ((b), (d)). The nonzero entries in $\w$ are complex Laplace distributed. Selected system parameter settings: $M = 100$, $N = 256$, and $K = 25$.\label{fig:laplace}}
\end{figure*}

We investigate the dependency of the performance of the considered algorithms on the underlying prior distribution of the non-zero entries in $\w$. To this end we repeat the previous numerical studies while considering two additional prior distributions for these entries. The first distribution results from selecting the nonzero entries to be of the form $\exp(j\phi_k)$, $k=1,\ldots,K$ with the  phases $\{\phi_k\}$ drawn independently and uniformly on the interval $[0,2\pi)$. The second distribution results from drawing the nonzero entries independently according to a complex Laplace distribution, see \eqref{eq:laplace_complex}, with unit variance. 
In the next comparison, the nonzero entries are iid according to the complex Laplace distribution with pdf \eqref{eq:laplace_complex} and variance one. We show results only for Fast-RVM, Fast-Laplace, and Fast-Besselk($\epsilon=0.5$), as the performance gain achieved by Fast-BesselK($\epsilon=0.5$) as compared to OMP and SpaRSA is similar to the performance observed in the previous investigations. We conclude from \Fig~\ref{fig:uniform} and \Fig~\ref{fig:laplace} that Fast-BesselK($\epsilon=0.5$) still maintains its superior performance. Furthermore, we again observe the important fact that Fast-BesselK($\epsilon=0.5$) achieves similar performance in scenarios with known or unknown noise precision. This is in direct contrast to the other Bayesian methods.  

\subsubsection{Performance for real signal models}

We conclude this section by briefly commenting on the performance achieved by the considered algorithms when they are devised for and applied to real-valued signal models. To distinguish between the algorithms devised based on real signal model from those devised for a complex signal model, in the subsequent discussion we refer to the former (latter) as real (complex) algorithms. 

In general, all considered complex algorithms perform better than their real variant. In particular, complex algorithms produce accurate results for less sparse weight vectors than their real counterpart. This is explained by the fact that the former use both real and imaginary parts to prune components in $\what$, thus, improving the sparse signal representation. 

The relative performances of the real algorithms compared to each other show the same trends as that observed for their complex variant. As an illustration, real Fast-BesselK($\epsilon=0$) is especially sensitive to high values of $K$; this is a well-known effect that arises when using the Jeffreys prior as the mixing density. This again emphasizes our conclusion that Fast-BesselK($\epsilon=0.5$) is a good trade-off between sparseness and reconstruction error.

\section{Conclusion \label{sec:conclusion}}
In this paper, we proposed a hierarchical prior model for sparse Bayesian learning (SBL) that applies to sparse signal representation in complex- and real-valued signal models. Our motivation was on the one hand to overcome the lack of sparsity-inducing prior models for complex signals and on the other hand to propose prior models that induce sparse, accurate, and robust signal representations in conditions of low and medium signal-to-noise ratio (SNR). Both aspects are of particular importance in many engineering applications of sparse signal representation, e.g., in wireless communications.

In the proposed hierarchical prior model the entries of the parameter vector of interest are modeled as independent complex Gaussian scale mixtures (GSMs) with mixing hyperparameters identically distributed according to a gamma distribution with shape parameter $\epsilon$ and rate parameter $\eta$. This model -- we termed it the Bessel K model -- comprises a family of hierarchical prior probability density functions (pdfs) indexed by these parameters. 

We analyzed the properties of \ti{} and \tii{} estimators derived from the Bessel K model. Our analysis revealed that the ability of a given element in the density family to induce sparse estimates heavily depends on the inference method used and, interestingly, whether real or complex signals are inferred. In the case of \ti{} estimation, the Bessel K model invokes, with the right setting of parameters $\epsilon$ and $\eta$, classical penalties such as the $\ell_1$-norm or the log-sum as special cases. The hierarchical Bayesian formulation of the $\ell_1$-norm penalty in the complex case is especially interesting as, to the authors' knowledge, it has not been proposed before. In the case of \tii{} estimation, the resulting penalties are also strongly influenced by the variance of the measurement noise, as pointed out by \cite{Wipf2011}. Nonetheless, we showed that the Bessel K model with $\epsilon<1$ promotes sparse \tii{} estimators even when the noise variance is high. In contrast, traditional prior models lose this property in such conditions. 

Finally, we derived a greedy algorithm of low complexity based on a modification of the expectation-maximization algorithm formulated for \tii{} estimation. As the Bessel K model encompasses as special cases previously proposed prior models, the algorithm generalizes existing fast SBL methods, allowing us to directly compare the impact of the different prior models on the performance of the resulting estimators.

The numerical results demonstrated that the Bessel K model with $\epsilon<1$ leads to estimators with superior convergence speed, sparseness, and lower mean-squared estimation error as compared to state-of-the-art sparse Bayesian estimators. We showed a significant robustness compared to the latter estimators in low and moderate SNR regimes. This is in agreement with the superior sparsity-inducing property of the Bessel K model with $\epsilon<1$ for highly noisy measurements, as shown in Section~\ref{sec:prior_models}. Furthermore, the results corroborate that the proposed estimators effectively include the estimation of the noise variance, thus avoiding the need for a training procedure for this parameter.

\appendix
\section{Approximate \ti{} Estimation Using EM \label{app:sbl_em1}}
Remember that the \ti{} estimator is the maximizer of 
\begin{align}
	\mathcal{L}(\w) 
= \log(p(\y|\w,\lambda)p(\w)).	
	\label{eq:sbl_cost1}
\end{align}
We formulate the EM algorithm approximating the \ti{} estimator by selecting $\{\gam,\y\}$ to be the complete data for $\w$. 
The E-step of the EM algorithm computes the conditional expectation
\begin{align}
  \langle \log p(\y,\w,\gam)\rangle_{p(\gam;\what)}
 \label{eq:completeloglikelihood1}
\end{align}
with 
\begin{align}
p(\gam;\what) \propto \prod_i \gamma_i^{\epsilon-\rho-1} \exp \big( -\gamma_i^{-1}\rho|\hat{w}_i|^2 - \gamma_i\eta \big)
\label{eq:qg}
\end{align}
computed in the E-step. 
The right-hand side expression in \eqref{eq:qg} is recognized as the product of GIG pdfs \cite{Joergensen}, i.e., $p(\gam)= \prod_{i} p(\gamma_i;\nu,a,b_i)$ where $p(\gamma_i;\nu,a,b_i)=\frac{(a/b_i)^{\frac{\nu}{2}}}{2K_{\nu}(\sqrt{ab_i})} \gamma_i^{\nu-1}\exp(-\frac{a}{2}\gamma_i -\frac{b_i}{2}\gamma_i^{-1})$ with order $\nu = \epsilon-\rho$ and parameters $a=2\eta$ and $b_i=2\rho|\hat{w}_i|^2$. The moments of the GIG distribution are given in closed form \cite{Joergensen}:
\begin{align}
\langle\gamma_i^n\rangle = \big(\frac{\rho|\hat{w}_i|^2}{\eta}\big)^{\frac{n}{2}}  \frac{ K_{\nu+n}(2\sqrt{\rho\eta}|\hat{w}_i|) }{K_{\nu}( 2\sqrt{\rho\eta}|\hat{w}_i|)}, \quad n\in \mathbb{R}. 
\label{eq:gamma_mean1}
\end{align}
The M-step of the EM algorithm updates the estimate of $\w$ as the maximizer of \eqref{eq:completeloglikelihood1}:
\begin{align}
	\what = \big( \Hmat^\hermit\Hmat + \lambda^{-1}\langle\Gammat^{-1}\rangle\big)^{-1}\Hmat^\hermit\y.
\end{align}
In case we use the Laplace GSM model ($\nu=\epsilon-\rho=1/2$), \eqref{eq:gamma_mean1} with $n=-1$ simplifies to  
\begin{align}
\langle\gamma_i^{-1}\rangle = \frac{\sqrt{\eta/\rho}}{|\hat{w}_i|},
\end{align}
where we have invoked the identity $K_\nu(\cdot) = K_{-\nu}(\cdot)$ \cite{Abramowitz}.
\section{Results for Section~\ref{subsec:fastscheme}}
This appendix contains the derivations of some results used in Section~\ref{subsec:fastscheme}.

\subsection{Computation of $\langle|w_i|^2\rangle$ \label{app:sbl_wi}}

We follow the approach in \cite{ShutinFastRVM} to compute $\langle|w_i|^2\rangle$. We can express $\langle|w_i|^2\rangle$ as $\langle|w_i|^2\rangle = \ev^\trans_i(\Sigmamat + \muv\muv^\hermit)\ev_i$ with $\ev_i$ being an $N\times1$ vector of all zeros with $1$ at the $i$th position. First, we consider the dependency of $\Sigmamat$ in \eqref{eq:wcov} on a single parameter $\gamma_i$. We note that $\Sigmamat=(\lambda\Hmat^\hermit\Hmat+\sum_{k\neq i}\gamma_k^{-1}\ev_k\ev_k^\trans+\gamma_i^{-1}\ev_i\ev_i^\trans)^{-1}$. Making use of the matrix inversion lemma \cite{Mardia1979} we recast $\Sigmamat$ as 
\begin{align} 
	\Sigmamat=\Sigmamat_{-i}-\frac{\Sigmamat_{-i}\ev_i\ev_i^\trans\Sigmamat_{-i}}{\gamma_i+\ev_i^\trans\Sigmamat_{-i}\ev_i},
\end{align}
where $\Sigmamat_{-i}\triangleq (\lambda \Hmat^\hermit\Hmat+\sum_{k\neq i}\gamma_k^{-1}\ev_k\ev_k^\trans)^{-1}$. After some straightforward algebraic manipulations,  $\langle|w_i|^2\rangle$ can be expressed as 
\begin{align}
\langle|w_i|^2\rangle = \frac{\gamma_i^2(s_i+|q_i|^2)+\gamma_is_i^2}{(\gamma_i+s_i)^2}
\label{eq:wiapp}
\end{align} 
with the definitions $s_i \triangleq \ev_i^\trans\Sigmamat_{-i}\ev_i$ and $q_i \triangleq \lambda\ev_i^\trans\Sigmamat_{-i}\Hmat^\hermit\y$.


\subsection{Computation of the stationary points of $\ell_i(\gamma_i)$ \label{app:sbl_ell}}
We define  
\begin{align}
\ell_i(\gamma_i) \propto^{e} \log(p(\y|\gamma_i,\gam_{-i},\lambda)p(\gamma_i)). \label{eq:ell_gammai}
\end{align}
Following the steps in \cite{Tipping2003} we can write $\ell(\gamma_i)$ as
\begin{align}
\ell_i(\gamma_i) \triangleq -\rho\log|1+\gamma_i\tilde{s}_i| + \rho \frac{|\tilde{q}_i|^2}{\gamma_i^{-1}+\tilde{s}_i} + (\epsilon-1) \log \gamma_i - \eta\gamma_i
	\label{eq:lgammaiapp}
\end{align}
with the definitions $\tilde{s}_i \triangleq \h_i^\hermit\matr{C}^{-1}_{-i}\h_i$, $\tilde{q}_i \triangleq \y^\hermit\matr{C}^{-1}_{-i}\h_i$, and $\matr{C}_{-i}\triangleq \lambda^{-1}\matr{I}+\sum_{k\neq i}\gamma_k\h_k\h_k^\hermit$. Taking the derivative of $\ell$ with respect to $\gamma_i$ and equating the result to zero yields 
\begin{align}
	0 &= \eta \tilde{s}_i^2\gamma_i^3 + \gamma_i^2[ 2\eta \tilde{s}_i - (\epsilon-\rho-1)\tilde{s}_i^2 ] + \gamma_i [ \eta +\rho(\tilde{s}_i - |\tilde{q}_i|^2)- 2(\epsilon-1)\tilde{s}_i ] - (\epsilon-1). 
	\label{eq:cubic2app}
\end{align} 
Making use of the matrix inversion lemma for $\matr{C}^{-1}_{-i}$, we show the identities $s_i = \tilde{s}^{-1}_i$ and $|q_i|^2 = |\tilde{q}_i|^2/\tilde{s}_i^2$ \cite{ShutinFastRVM}. By substituting these identities into \eqref{eq:cubic}, we arrive at the cubic equation in \eqref{eq:cubic2app}. Thus, the positive solutions of \eqref{eq:cubic} are the stationary points of \eqref{eq:lgammaiapp}. 

\section*{Acknowledgment}
This work was supported by the 4GMCT cooperative research project funded by Intel Mobile Communications, Agilent Technologies, Aalborg University and the Danish National Advanced Technology Foundation.


\bibliographystyle{elsarticle-num} 
\bibliography{mybib}


%
%
%
%
\end{document}